\pdfoutput=1
\documentclass[11pt]{article}
\usepackage[a4paper, bindingoffset=0.2in, left=1in, right=1in, top=1in, bottom=1in, footskip=.25in]{geometry}
\usepackage{subcaption}
\usepackage{booktabs}
\usepackage{multirow}
\usepackage{setspace}
\usepackage{url}
\usepackage{authblk}
\usepackage{tikz}
\usepackage{nicematrix}
\usepackage{amsmath}
\usetikzlibrary{matrix,backgrounds}
\usepackage{hyperref}

\usepackage[style=apa,natbib=true, url=false, doi=false, uniquename=false]{biblatex}
\AtEveryBibitem{%
  \clearfield{issn} 
  \clearfield{note} 
  \clearfield{doi} 
  \clearfield{urldate}
  \ifentrytype{online}{}{
    \clearfield{url}
  }
}
\usepackage{csquotes}

\title{Improving Computer Vision Interpretability: Transparent Two-level Classification for Complex Scenes}
\author[1]{Stefan Scholz \thanks{Corresponding author. Email: stefan.scholz@uni-konstanz.de}} 
\author[1]{Nils B.~Weidmann} 
\author[2]{Zachary C.~Steinert-Threlkeld} 
\author[1]{Eda Keremo\u{g}lu} 
\author[1]{Bastian Goldlücke} 
\affil[1]{Center for Image Analysis in the Social Sciences, University of Konstanz}
\affil[2]{Luskin School of Public Affairs, University of California, Los Angeles}
\date{July 4, 2024}

\begin{document}

\maketitle

\onehalfspacing

\begin{abstract}
    Treating images as data has become increasingly popular in political science. While existing classifiers for images reach high levels of accuracy, it is difficult to systematically assess the visual features on which they base their classification. This paper presents a two-level classification method that addresses this transparency problem. At the first stage, an image segmenter detects the objects present in the image and a feature vector is created from those objects. In the second stage, this feature vector is used as input for standard machine learning classifiers to discriminate between images. We apply this method to a new dataset of more than 140,000 images to detect which ones display political protest. This analysis demonstrates three advantages to this paper's approach. First, identifying objects in images improves transparency by providing human-understandable labels for the objects shown on an image. Second, knowing these objects enables analysis of which distinguish protest images from non-protest ones. Third, comparing the importance of objects across countries reveals how protest behavior varies. These insights are not available using conventional computer vision classifiers and provide new opportunities for comparative research.   
\end{abstract}

{\vspace{5mm}\centering\small\textbf{Keywords:} Image analysis, computer vision, explainable AI, two-level classification, protest analysis}

\clearpage

\section{Introduction}

Recent progress in the field of artificial intelligence and computer vision has led to an increasing adoption of image analysis in the social sciences. Images have a number of advantages over textual sources. They are language agnostic, so one can train one model instead of one model per language. They can also improve the measurement of concepts that are typically not mentioned in text, such as violent tactics, crowd composition, or the use of symbols \citep{abrams2023symbolic}. These advantages have led to innovative work in political science that studies, for example, the emotional impetus of images \citep{Casas2019imagesBLM}, altered vote tally sheets \citep{cantu2019fingerprints}, or media coverage of politicians \citep{girbau2023face}.

Currently, almost all computational image analysis is performed using deep neural networks. While these networks are able to achieve an impressive level of accuracy, it is difficult for the researcher to understand \textit{why} they assign a particular image to a given label or category. This problem is especially pressing when the image is \emph{complex}: the classification of an image that contains many different types of objects is more difficult to understand than the classification of simple images showing single persons, maps, or ballots. As these methods continue to grow in importance in a wide range of research, the necessity of interpreting their operation has become a growing area of research \citep{rudin2019stop}. This paper presents an approach that helps remedy the opacity of vision models, such that they can be explored further for social science applications.

The paper introduces a two-level image classification method to improve computer vision interpretability. First, one creates a feature vector from the objects (``segments'' in computer vision terminology) an image contains. Next, a non-visual machine learning classifier uses the feature vectors to identify combinations of objects predictive of the researchers' outcome of interest. We demonstrate this approach on a new dataset of 141,538 protest images from ten countries. Mass protest is an increasingly frequent phenomenon: whether the issue is COVID-19 lockdowns in China, womens' rights in Iran, or indigenous rights in Peru, their global number has increased steeply in recent years \citep{ortiz2022world}. The rise of social media has led to a concurrent profusion of images documenting protest, and researchers have started to build protest event datasets from them \citep{zhang2019casm,steinert-threlkeld2022how}. Protest images are instances of complex scenes mentioned above: they are frequently composed of different objects such as people, flags, signs, or cars.

This example demonstrates three advantages of the approach. First, the two-level classifier identifies the objects an image contains, allowing for immediate understanding of image content unlike existing pixel-based methods. These objects are human-understandable items such as ``car" or ``fence", an improvement over approaches that identify pixel activations of areas of high contrast pixels \citep{torres2023bovw}. Second, permutation tests reveal which objects distinguish protest from non-protest images, which allow for simple validation tests of the classification. Third, the distribution of object permutation importance across protests shows how symbolic usages varies across different national and political contexts. None of these insights are available using current image classifiers and the pixel-based interpretation techniques applied to them. Researchers seeking to apply our method to their own images can do so with our pre-configured online demonstration tool and the associated API for batch processing (see data availability statement).

\section{Image Classification: The Limited Interpretability of Conventional Approaches}

Prior to the introduction of multilayered (``deep'') neural networks, computer vision used less flexible statistical models such as color distributions or predefined feature maps. The introduction of deep neural networks, the explosion of training dataset size, and use of specialized hardware has brought forth models that assign proper labels, recognize objects, and analyze faces markedly better than previous methods \citep{lecun2015deep}. 

A convolutional neural network (CNN) consists of a series of layers, ranging from feature extraction to the fully connected layers. The last of these layers then feeds into a function that outputs a vector as long as the number of classes. The output of AlexNet, for example, is 1,000 units long because the dataset it trained on contains 1,000 labels such as dog, cloud, and car \citep{krizhevsky2012imagenet}. In contrast, the output of the model in \citet{cantu2019fingerprints} is two units long because the labels are ``altered ballot'' and ``not altered ballot.'' This final vector, the output layer, is the model's estimate of the content of an image. See \citet{casas2020image}, \citet{joo2022image}, and \citet{torres2022learning} for more on the functioning of CNNs.

The parameters estimated during training are the values of the kernels and the weights connecting neurons in the fully connected layers. Deep neural networks' hierarchical structure, numerous feature maps, and use of fully connected layers mean they contain tens of millions of parameters. This complexity makes it impossible to determine \textit{why an image receives the classification it does} by looking at parameter values, a marked contrast to regression models' coefficients or simpler machine learning classifiers' parameters, e.g.~trees in a random forest.

Different methods have been developed to aid the interpretation of computer vision models. Since pixels are the atomic elements of images, like characters in text, it is natural to ask which pixels contribute to a model's classification of an image. An early example of this approach is deconvolutional neural networks (``DeconvNets''). A DeconvNet estimates important pixels from feature maps \citep{zeiler2014visualizing}. For a given feature map and layer, the DeconvNet can show which pixels, such as those associated with a door or face, most activate that feature. The output is the input image at that layer with the non-activated pixels removed. Another approach is gradient class-activation mapping \citep[Grad-CAM,][]{selvaraju2017gradcam}. This method sets the class score to 1 for the desired class, e.g. ``protest," and backpropagates the activation to the pixels in a given image driving that classification; the output is the input image with a pixel heatmap.

A third approach is integrated gradients: the classified image is compared to a baseline image, usually a black square, and each pixel's contribution to the final class score is compared to its prediction when the baseline image is assigned to that class \citep{sundararajan2017axiomatic}. The output is the original image with pixels colored by their gradient sum. A fourth approach is attention maps, which are calculated using the attention mechanism of transformer models \citep{dosovitskiy2020image}. The attention mechanism of vision transformers (ViTs) allows them to focus on specific areas of images, as opposed to convolutional neural networks (CNNs). The attention weights can be visualized post hoc in attention maps, where the higher the attention weight, the brighter the color of the pixel in the attention map.

\begin{figure}[t!]
    \centering
    \begin{subfigure}{0.33\textwidth}
        \caption{Deconvolution}
        \includegraphics[width=\textwidth]{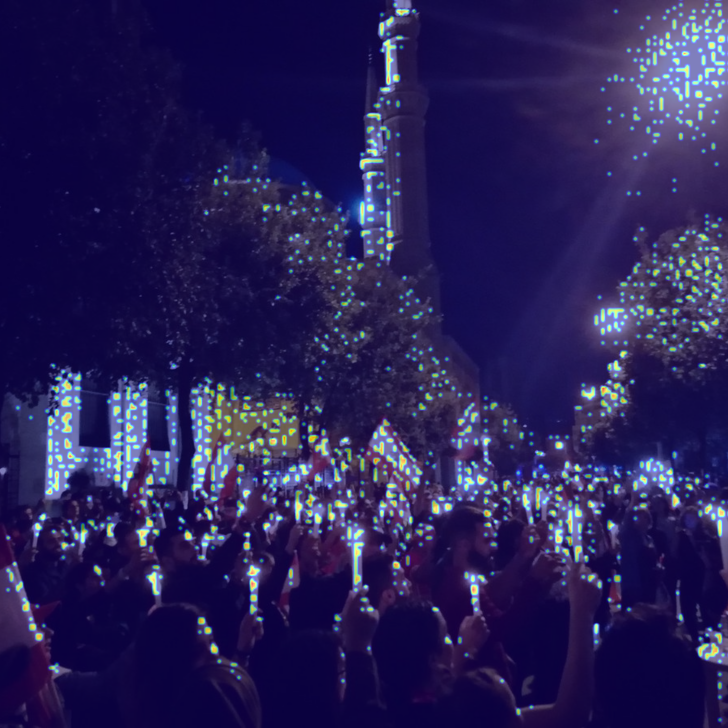}
    \end{subfigure}
    \hspace{2em}
    \begin{subfigure}{0.33\textwidth}
        \caption{Grad-CAM}
        \includegraphics[width=\textwidth]{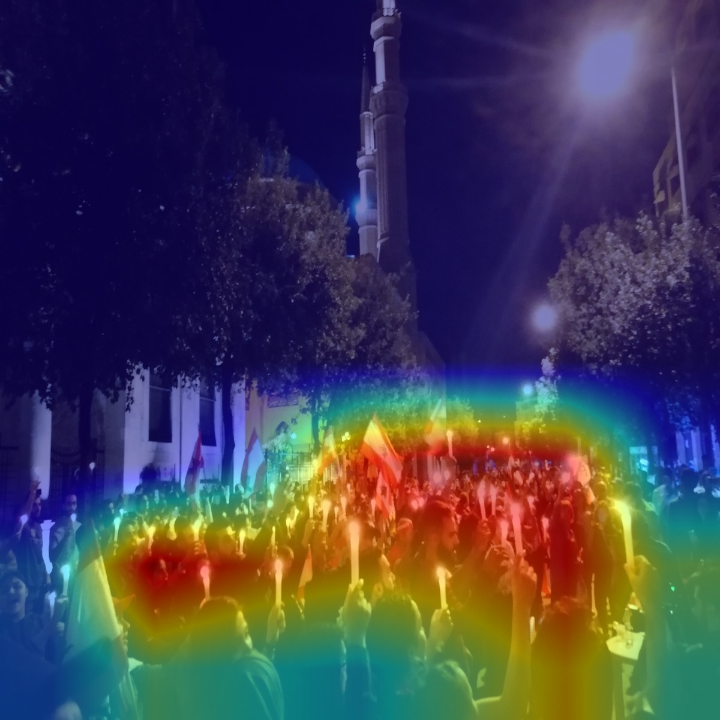}
    \end{subfigure}
    \bigskip
    \begin{subfigure}{0.33\textwidth}
        \caption{Integrated Gradients}
        \includegraphics[width=\textwidth]{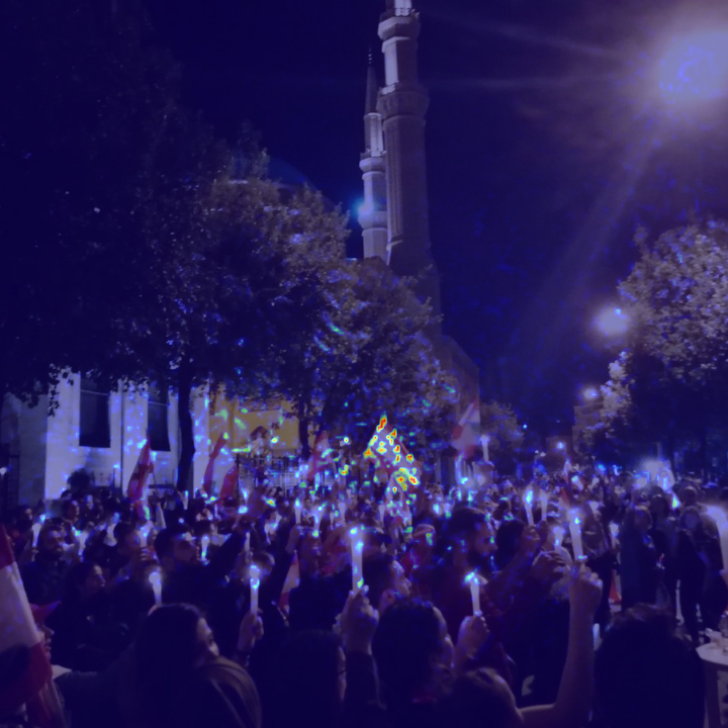}
    \end{subfigure}
    \hspace{2em}
    \begin{subfigure}{0.33\textwidth}
        \caption{Attention Rollout}
        \includegraphics[width=\textwidth]{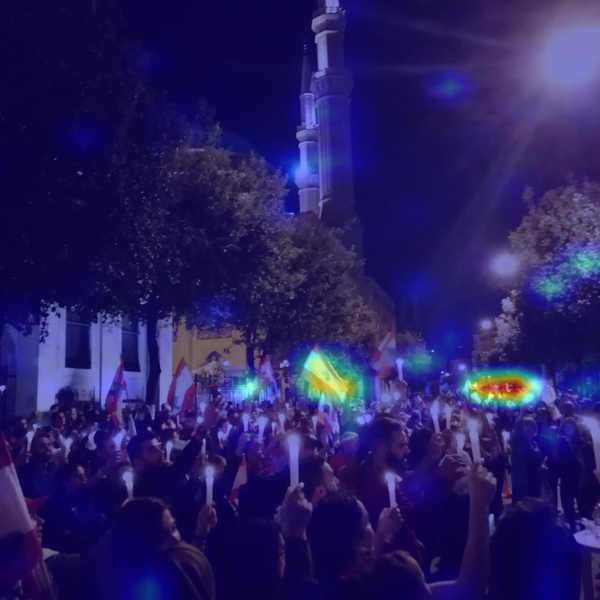}
    \end{subfigure}
    \caption[Comparison of visual information extracted from protest image]{Comparison of visual information extracted from protest image with Deconvolution, Grad-CAM, Integrated Gradients and Attention Rollout.}
    \label{fig:visual_comparison}
\end{figure}

Figure \ref{fig:visual_comparison} shows a protest image with the four pixel-based visual explanations described above.\footnote{As computer vision models we use the ResNet50 to visualize deconvolution, Grad-CAM and integrated gradients, and vision transformer (ViT) models to visualize attention rollout. Both were self-trained on our dataset of protest images.} In each case, pixels are colored as a heatmap based on their relative contribution to classifying the image as containing a protest. While the highlighted areas provide plausible cues as to why the image is classified as a protest image, Figure \ref{fig:visual_comparison} reveals two major shortcomings. First, though important pixels are highlighted, the researcher still must determine the concept or object behind a collection of pixels. For example, in Panel (b) of Figure \ref{fig:visual_comparison}, it is unclear if the protest classification is driven by a group of people, their attire, or the presence of flags. Second, different visual explanation approaches emphasize different parts of the image. Comparing Panels (b) and (d), for example, the latter suggests that classification is driven by different, smaller parts of the image rather than the group of people in the former: a flag, two indecipherable areas on each side of the flag, and even the foliage of a tree.

For comparative research with images, these difficulties pose major obstacles. First, coding criteria should be explicit, such that it is possible to understand why a particular category or label has been assigned. This is a requirement regardless of how the coding was done (human or automatic). In automatic text classification, for example, it is possible to identify words or word combinations that pertain to particular topics, which allows researchers to check the validity of the coding. In image classification, however, there is no natural unit into which images can be decomposed, leading to a second obstacle. Absent an abstract description of the content of an image, it is difficult to compare images across context and cases to see whether and why their content differs. 

The method proposed in the next section obviates these problems by identifying specific objects in images (first level) and then using those objects to determine if an image contains the desired concept (second level). Instead of looking at individual pixels, it looks at objects (groups of pixels) in an image. It then uses the differential presence of certain classes of objects to determine whether an image contains the desired concept (for this paper, protest). This approach is different to pixel-alignment methods because it does not need to explain which pixels are aligned with which objects. 

\section{Two-level Classification}
\label{sec:method}

This paper introduces a two-level process for image classification. The first level creates a vector summarizing the objects contained in each image; the second trains a non-visual classifier on these vectors. This section describes each step in more detail.

\subsection{Creating Feature Vectors from Images}

The first step maps the pixel representation of images to objects, an interpretable lower dimensional representation. We first describe how to extract the objects from an image, before describing how to turn them into vectors that summarize the objects contained in them.

\subsubsection*{Extracting Segments}

In addition to classifying entire images, computer vision models can detect and classify \textit{objects within images}. Object detection refers to estimating bounding boxes around potential objects and classifying these areas into objects of different categories. An extension of object detection is instance segmentation, where rather than simple bounding boxes, pixel masks are provided for the shapes of the detected objects. In recent years, a number of frameworks have been proposed which have increased the accuracy and efficiency of instance segmentation \citep[e.g.][]{girshick2014rich, he2017mask}. The most commonly used metric to measure the accuracy is average precision (AP). It rewards correct classifications and precise masks, which means that the higher the AP, the better the framework. These frameworks are trained in a fully supervised fashion on particular datasets and can therefore be readily applied to new images. 

The datasets used for training segmentation algorithms define what types of objects these algorithms can later detect. The entire set of these object types (or categories) is called the ``vocabulary'' of the segmenter. Older approaches such as the \textit{PASCAL Visual Object Classes} \citep[PASCAL VOC,][]{everingham2010pascal} have small vocabularies (20). More recent ones, such as the \textit{Common Objects in Context} dataset \citep[COCO,][]{lin2014microsoft} has 80 categories, and the \textit{Large Vocabulary Instance Segmentation} dataset \citep[LVIS,][]{gupta2019lvis} has 1,203; these two datasets are developed on the same set of images. The rest of the paper focuses on the COCO and LVIS datasets. 

\begin{figure}[b!]
    \centering
    \begin{subfigure}{0.32\textwidth}
        \includegraphics[width=\textwidth]{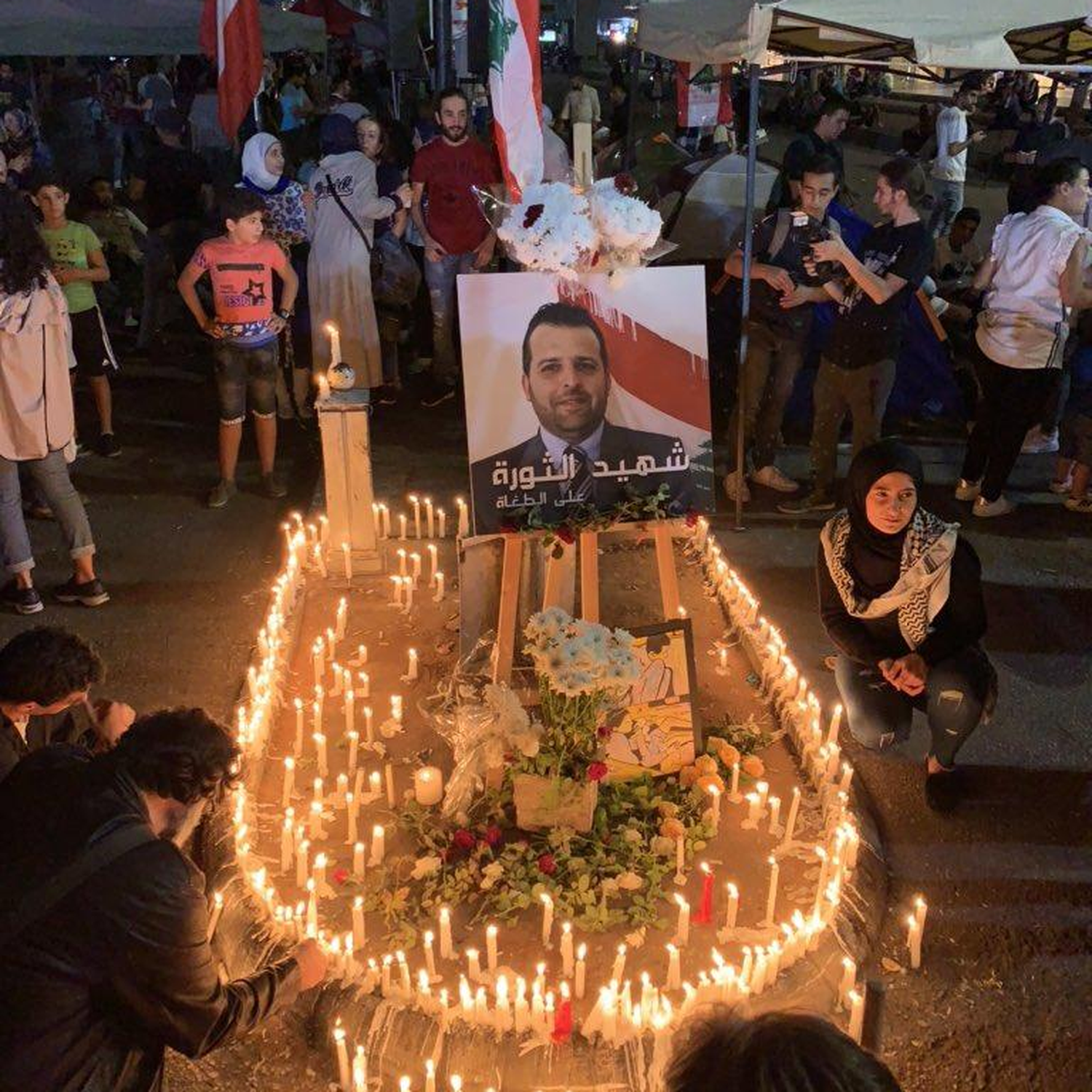}
    \end{subfigure}
    \begin{subfigure}{0.32\textwidth}
        \includegraphics[width=\textwidth]{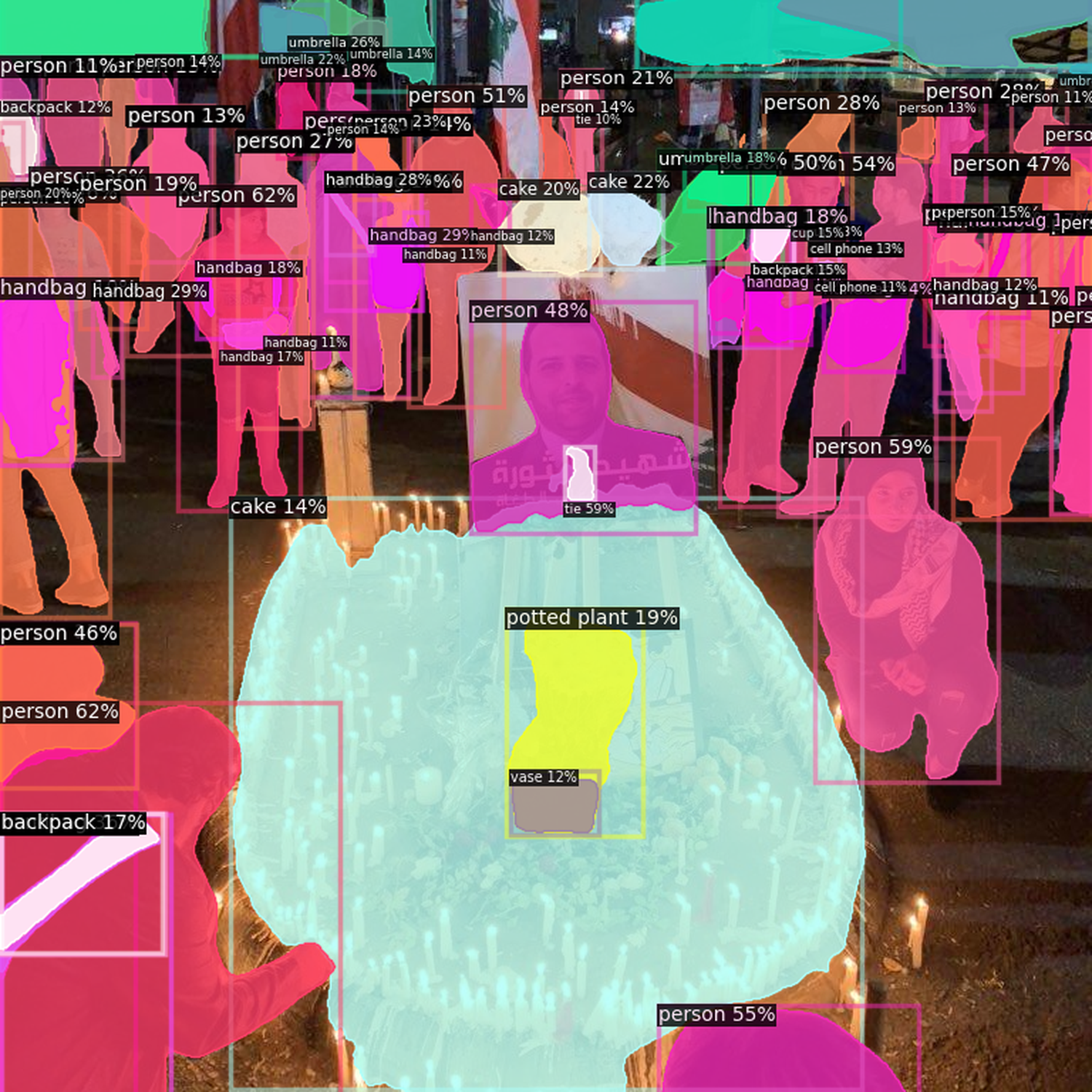}
    \end{subfigure}
    \begin{subfigure}{0.32\textwidth}
        \includegraphics[width=\textwidth]{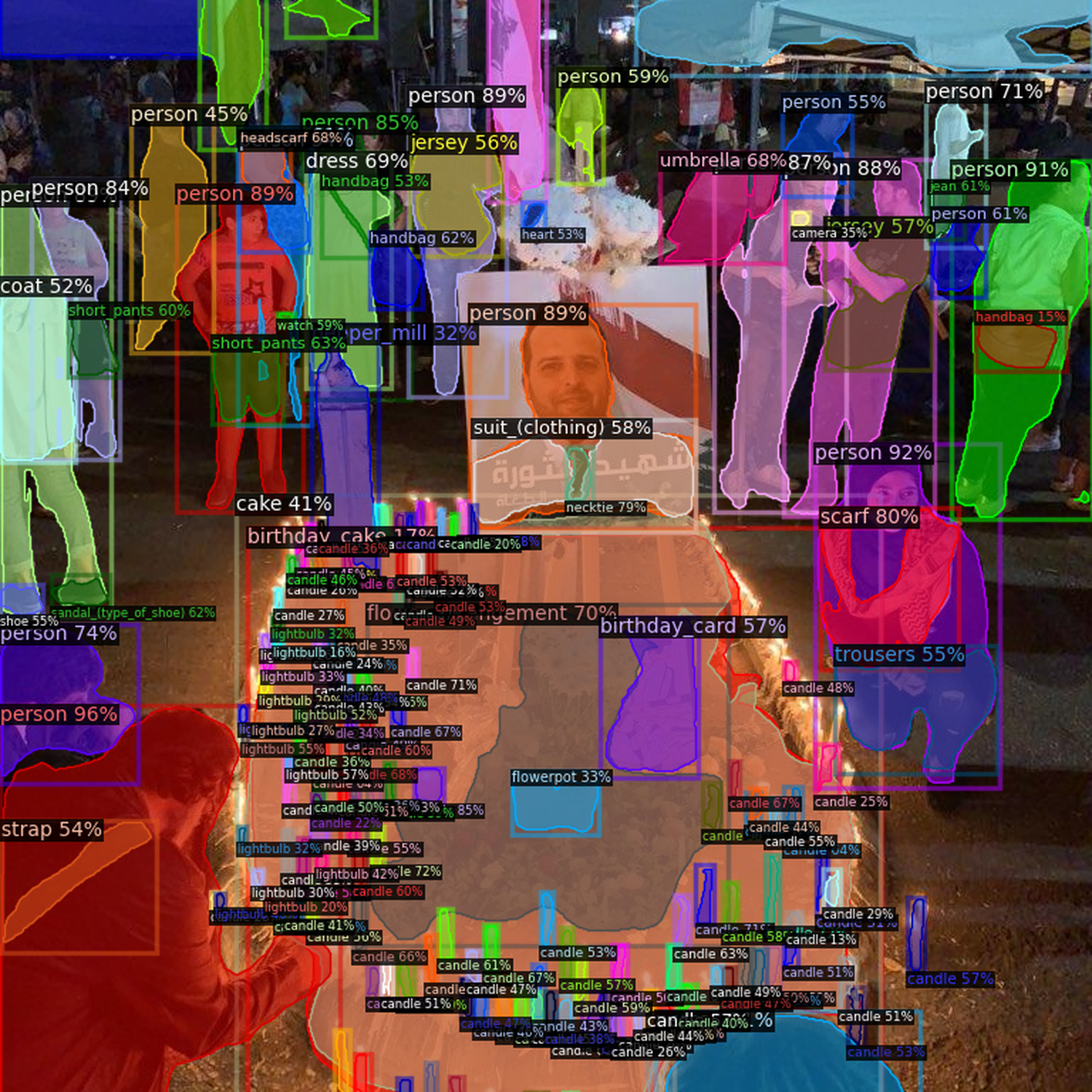}
    \end{subfigure}
    \caption[Vocabularies]{Instance segmentation applied to an image of a candlelight vigil (left) using COCO vocabulary (center) and LVIS vocabulary (right).} 
    \label{fig:segmentation-example}
\end{figure}

Figure \ref{fig:segmentation-example} shows the segments detected in a protest picture of individuals at a vigil with candlelights. The center panel of Figure \ref{fig:segmentation-example} uses an instance segmentation method trained using the COCO vocabulary, and the right panel shows segments detected with the LVIS vocabulary. Not surprisingly, LVIS detects more segments and more segments of different object types, providing greater detail about the image content than COCO.

\subsubsection*{Creating Segment Vectors}

All instance segmenters generate object positions, categories, and confidence scores for each detected segment of the input image. The position is called the mask; it is a polygon outlining the proposed object. Since the segmentation method cannot always assign a unique predicted object category, it outputs certainty scores (0-1) for each category in the vocabulary on which the detector was trained. Following conventional approaches \citep[e.g.][]{he2017mask}, for each segment, we assign the object category with the highest certainty score to the respective segment. Then, we use this information to create abstract descriptions of the set of objects contained in an image. The collection of objects is used to create a feature vector.

\begin{figure}[b!]
    \centering
    \begin{subfigure}[t]{0.48\textwidth}
        \centering
        \includegraphics[height=52mm]{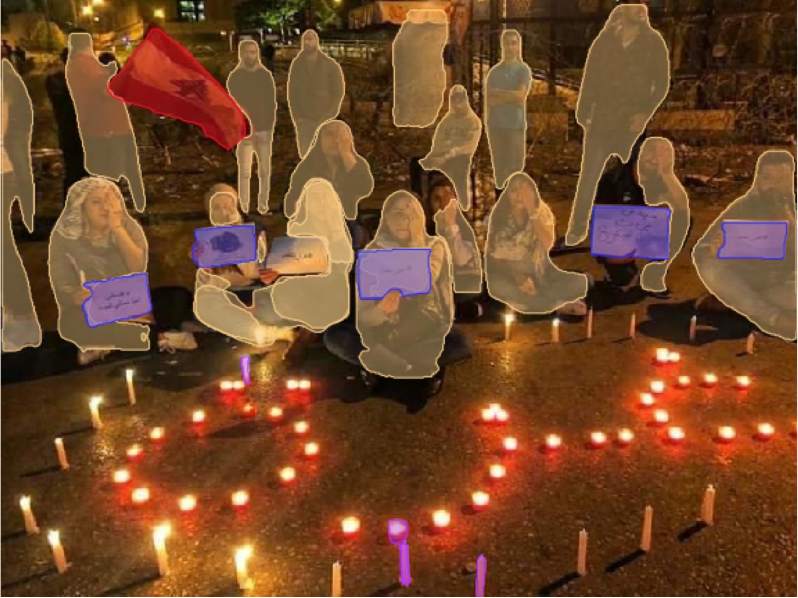}
    \end{subfigure}
    \begin{subfigure}[t]{0.48\textwidth}
        \centering
        \small
        \def\arraystretch{1.5}
        \raisebox{25mm}{
        $
        \begin{NiceMatrix}%
          [ 
            first-row,
            first-col,
            nullify-dots,
            cell-space-limits=1pt
          ]
         & \textrm{Poster} & \textrm{Person} & \textrm{Flag} & \textrm{Candle} & \textrm{Gun} \\[0.3cm]
         v_{a} & 1 & 1 & 1 & 1 & 0 \\[0.3cm]
         v_{b} & 5 & 19 & 1 & 4 & 0 \\[0.3cm]
         v_{c} & 0.01 & 0.03 & 0.02 & 0.00 & 0.00 \\[0.3cm]
         v_{d} & 0.03 & 0.28 & 0.02 & 0.00 & 0.00 \\[0.3cm]
        \CodeAfter
          \SubMatrix[{1-1}{1-5}]
          \SubMatrix[{2-1}{2-5}]
          \SubMatrix[{3-1}{3-5}]
          \SubMatrix[{4-1}{4-5}]
        \end{NiceMatrix}
        $}
        \normalsize
    \end{subfigure}
    \caption[Feature generation]{Feature generation from a segmented image (left), with different feature vectors generated from this image (right): binary vector ($v_a$), count-based vector ($v_b$), area-max vector ($v_c$), and area-sum vector ($v_d$).}
    \label{fig:feature-generation}
\end{figure}

There are a large number of ways to transform the output from the segmentation method into a feature vector. Figure \ref{fig:feature-generation} presents the four this paper evaluates. Each entry in the generated vectors corresponds to one type of object from the vocabulary of the segmentation method.

\paragraph{Binary features} A binary feature vector indicates the presence or absence of a certain object category in the image. The top feature vector $v_a$ in Figure \ref{fig:feature-generation} shows this construction for five object types.

\paragraph{Count-based features} An extension of the binary feature is to count how many objects of each object category are present. The second vector $v_b$ in Figure \ref{fig:feature-generation} illustrates this approach. The model detects five segments with a poster, 19 segments with a person, one segment with a flag, four segments with a candle, and no segments with a gun.

\paragraph{Area-based features} The positional information obtained from the segmenter can also be incorporated into the feature vectors. We do this in two ways. A third type of feature vector uses the maximum area of any object of a given category in the image, assuming that bigger objects are more important. In Figure \ref{fig:feature-generation}, vector $v_c$ indicates that the largest person on the image occupies 3\% of the image area. A fourth feature type uses the sum of the areas identified for each object category and therefore captures what proportion of the entire image objects of each type occupy. The bottom vector $v_d$ in Figure \ref{fig:feature-generation}, for example, indicates that persons take up 28\% of the image.

\subsection{Classification}

At the second level, we train a standard machine learning classifier to predict the image labels (protest) from the segment vectors of the images. We rely on four different classifiers: logistic regression, simple decision trees, collections of decision trees, and gradient-boosted decision trees. Logistic regression was chosen since it is widely used by social scientists, and to provide a benchmark against. The tree-based classifiers were selected because they allow the researcher to vary complexity and interpretability.

The simplest decision tree can have a depth of two and classify an observation using one condition. Decision trees can become more complex when the depth of the tree -- the number of conditions -- is increased. Combining decision trees into an ensemble allows each tree to attempt to correct the misclassifications of its predecessor tree. These gradient-boosted decision trees improve accuracy, though the added complexity inhibits interpretability because any observation now follows multiple trees. In practice, the number of trees -- the number of boosting rounds -- can reach into the thousands.

\section{Application: Coding Protest Images}

This section develops and evaluates a two-level classifier using a new dataset of protest images.

\subsection{A New Protest Image Dataset}
\label{sec:data}

A new dataset of protest images collected from social media is used to test the performance and added value of the two-level classifier. This dataset focuses on large protest episodes in different countries worldwide. To build the dataset, the \textit{Armed Conflict, Location, and Events Dataset} \citep[ACLED,][]{raleigh2010introducing} was used to identify 46 country-periods with many protests from 2014-2021 that are also high-income and populous enough to generate enough social media reporting \citep{steinert-threlkeld2022mmchived}. These 46 country-periods were narrowed to 14 based on their Polity IV score and region, with a goal of generating broad coverage of regime types and parts of the world. Twitter is the platform used to obtain images because of its widespread use. This step generates 135 million tweets and 16 million images. 

The number of tweets a country produces is a function of a country's population, its gross domestic product, and the duration of a protest. There is therefore a large difference in the number of images per country. To avoid any potential bias, we downsample each country's images to 100,000. For the three countries that have fewer than 100,000 images, all are kept. Appendix A.1 presents more details on the selection of countries and images.

A protest is defined as a publicly visible event or action involving at least one person making a political statement or expression. Human coders assigned one of four labels: protest (with high certainty), protest (low certainty), no protest (low certainty), or no protest (high certainty). The use of the different degrees of certainty captures the fact that oftentimes images alone lack the context of a larger protest episode to be interpreted with certainty.\footnote{While other work improved classification by combining text and image data \citep{zhang2019casm}, our procedure emulates a scenario where the coding is based on visual material alone.} Exact and near duplicate images are removed before labeling (see Appendix A.2). For more details of the coding procedure, see Appendix A.3 and A.4.

Once labeled, images from Japan, Kazakhstan, Ethiopia, and the Philippines are also excluded because there were fewer than 100 protest images. The remaining 141,538 images are split into 80-20 training and testing sets by country (see Appendix A.5 for details). In the training dataset, there are 12,454 protest images and 100,776 non-protest images. In the testing dataset, 3,113 and 25,195 respectively. This relative paucity of protest images matches the distribution found in other work \citep{steinert-threlkeld2022how}.

\subsection{Training the Two-Level Classifier}
\label{sec:trainingCV}

Using the new protest image dataset, we construct our two-level classifier using different combinations of vocabularies, feature vectors and second-level classifiers. 

\paragraph{Vocabulary} We use the COCO and LVIS vocabularies introduced earlier. We detect the segments in the COCO vocabulary \citep{lin2014microsoft} with the instance segmentation model by \citet{li2023mask}. This model was validated on the COCO dataset with a mask average precision (AP) of 0.5230. To detect the segments in the LVIS vocabulary \citep{gupta2019lvis}, we use the instance segmentation model by \citet{zhou2022detecting}. This model achieved a mask AP of 0.2497 on the LVIS v1.0 validation set, which is not the highest AP reported in the original paper, but visually has the best results and more reliably recognizes frequent categories. Segments with a confidence score below 0.1 are discarded because visual inspection shows segments below this threshold are not reliable. 

\paragraph{Feature generation} To explore how different kinds of feature vectors affect the classifier performance, the four types of vectors introduced earlier are used: binary features (at least one segment of category on image), count-based features (sum up the number of segments of category on image), the maximum-area features (the area occupied by the largest instance of a category on an image), and the sum-area features (the total area occupied by all instances of a category on the image).

\paragraph{Classification method} For classification of the feature vectors, four different classification methods are used at the second level: A logistic regression, a simple decision tree, a random forest \citep{breiman2001random} and an XGBoost gradient-boosted tree \citep{chen2016xgboost}. These classification methods have different hyperparameters that must be chosen. We optimize them using 5-fold cross-validation on the training set; more details are reported in Appendix B.

\section{Results}
\label{sec:results}

This section evaluates how the different design choices of the classifier affect its performance. We then present three sets of results: comparing the two-level classifier to existing conventional computer vision approaches, showing which objects are most important for identifying protest images, and using the by-country importance to understand protests in a comparative context. All results shown here are based on the dataset described above, which combines low and high confidence labels. Section E in the Appendix shows the results of the subsequent analysis with only high certainty images. 

\subsection{Design Choices and Classification Performance}

\begin{figure}[b!]
    \centering
    \includegraphics[width=0.96\textwidth]{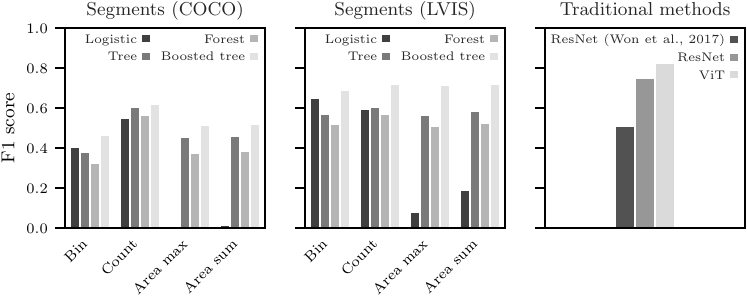}
    \caption{Out-of-sample evaluation of different methods. The figure displays the F1 score achieved on the test set; LVIS area sum with gradient-boosted trees achieves the best F1 score of 0.7203. The logistic regression obtains low F1 scores with the area-based features, making certain bars invisible. Visualization based on Table A3 in the Appendix.}
    \label{fig:models}
\end{figure}

The full combinatorial space of vocabularies, feature vectors, and second-level classifiers is explored to determine the best two-level classifier, where performance is measured using F1 scores, a commonly used metric suitable for imbalanced class distributions. The left two panels of Figure \ref{fig:models} show the results. How do the vocabulary, the types of feature vectors, and the second-level classifier affect predictive performance?

\paragraph{Vocabulary} Comparing models trained on the COCO vocabulary and those trained on the LVIS vocabulary (left and center panel of Figure \ref{fig:models}) shows that the latter mostly achieve better results. With the 80 objects categories available from the COCO vocabulary, F1 scores rarely exceed 0.5, while those relying on LVIS achieve F1 scores of up to 0.7203. These results show that the performance for most second-stage models and feature generation methods is improved by incorporating more object categories.

\paragraph{Feature generation} Feature generation affects the second-stage classifier's performance, though the difference within vocabularies is less than that across them. There are two exceptions. First, count-based features perform particularly well among the models using the COCO vocabulary and second, logistic models perform poorly for the area feature vectors with the LVIS vocabulary. In all other cases, the creation of the feature vectors does not seem to play a major role. This is not too surprising, since they all encode the presence/absence of different object categories in different ways, which seems to be sufficient for classification. 

\paragraph{Classification method} Logistic regression achieves its best results with the binary and count based features, but their results deteriorate significantly with the area based features. The random forest based models do not improve the accuracy compared to the logistic regression models for the binary and count feature vectors, but they do for the two area ones. Gradient-boosted trees generally improve the F1 score, in particular for the larger vocabulary. This result matches other work that has found random forests and gradient-boosted trees to outperform logistic regression in classifying rare events \citep{muchlinski2016comparing,wang2019comparing}.

We also conduct a more in-depth analysis of our classifier to see if it systematically misses images with particular motives or content. To this end, we cluster the images into 30 topics and report the confusion matrix, precision, recall, and F1 of the classifier for each topic separately. Appendix D describes the clustering process in detail and presents the performance of the LVIS area-sum gradient-boosted classifier in these clusters. In the cluster of images showing protests, fires and smoke, the classification performance is below average, as it is for images with state police. Protests with large crowds, individual banners or flags, on the other hand, are classified more accurately. However, we do not see a particularly poor performance for some clusters, which suggests that the classifier does not systematically fail to discover certain kinds of protest images.

\subsection{Comparison to Conventional Approaches}

To compare the two-level classifier to other computer vision approaches, three other models are used. Relying on earlier work in this area, we use the ResNet50 from \citet{won2017protest}, a convolutional neural network (CNN) trained on a dataset of more than 40,000 online images. The second is the same ResNet50 CNN but trained on this paper's protest image dataset. A larger CNN such as a ResNet101 or ResNet152 could further improve accuracy, but a ResNet50 with the same training times and hardware requirements is used to facilitate comparison. The third model is a vision transformer (ViT), a model that has been shown to outperform CNNs on many computer vision tasks while requiring less computational resources \citep{dosovitskiy2020image}. We also use the smaller ViT version (``base''), whose entire weights are also learned during several epochs of training (see Appendix B for details on model training). 

Figure \ref{fig:models} (right panel) shows the results. The ResNet50 from \citet{won2017protest} achieves the worst fit of the classifiers, most likely because it was trained on a different, smaller dataset. Training a ResNet50 on our dataset generates an F1 score of 0.7489 on the test data, almost identical to the F1 score of the best segments-based classifier. The ViT model achieves an F1 score of 0.8226 on the test data, and clearly outperforms our two-level classification. Reducing images to identifiable objects may therefore lose information that improves classification. This can happen because some objects that are relevant for protest fail to be included in the generic vocabularies used or because context of objects in the images is lost by extracting the segments only. Nevertheless, while the predictive performance of the segment-based classifier is not as strong as the cutting-edge in computer vision, its performance remains comparable to other established methods. At the same time, it provides different new possibilities, as we demonstrate in the following sections. 

\subsection{Importance of Features in Protest Images}

\begin{figure}[b!]
    \centering
    \includegraphics[width=0.96\textwidth]{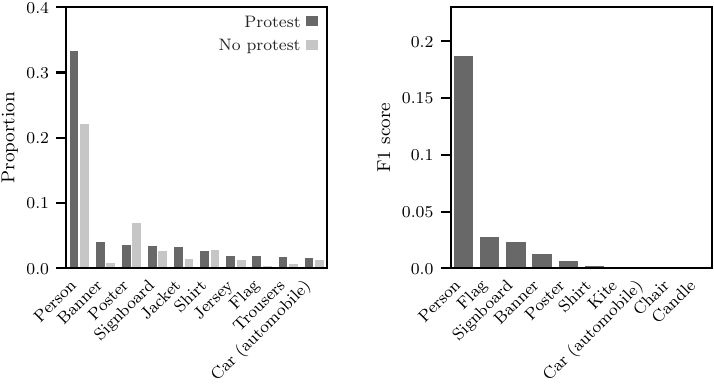}
    \caption{Proportion of segments in protest and non-protest images (left) and importance of area-sum aggregated segments (right).} 
    \label{fig:features-similarities}
\end{figure}

The two-level classifier's advantage is that, by design, it offers insights into which object categories are related to protest. In a first illustration of this advantage, we examine which types of objects are most likely to appear in protest images. Figure \ref{fig:features-similarities} presents the total area occupied by instances from a certain segment category in protest and non-protest images when using the LVIS area-sum gradient-boosted tree classifier. The proportions are calculated for each country separately and then averaged. For clarity the analysis is limited to the ten largest segment categories of protest images. Not surprisingly, the largest segments in a protest image are persons: 34\% of a protest image is occupied by people versus 22\% for non-protest images. Protest images also largely display banners, posters, signboards, jackets, shirts, jerseys, flags, trousers, and cars. It is not always the case that protest images seem to be characterized by larger total areas of these objects types, however. For example, while posters are frequently shown on protest images, non-protest images tend to have larger areas occupied by posters. The reason may be that the presence of a poster alone (as for example, in a close-up photo) is not sufficient for an image to be classified as a protest image.

However, the largest categories are not necessarily the most important ones for prediction. To find out how essential an object category is for classifying protest, we calculate the importance of the corresponding features. For this purpose, a feature is randomly permuted to break its relationship to the labels. This random permutation is repeated for each feature, always followed by a reevaluation of the F1 score of the classifier on the training set. Thus, if this random permutation leads to a higher drop in performance for a given feature, this object category is more essential for the model to distinguish a protest image from a non-protest image. Rather than determining the feature importance for all images of all countries at the same time, the feature importance is calculated for each country separately and then averaged. In this way we evaluate which features seem to be similarly important across all countries irrespective of the number of protest images we have for them.

The right panel in Figure \ref{fig:features-similarities} presents the results of this importance test. The model has the largest drop in F1 when the total area occupied by persons is permuted (0.19); followed by flag (0.03), signboard (0.02), and banner (0.01). As expected, almost all of the object types shown in the plot are closely related to protest, confirming the validity of the paper's method. In addition, the largest object categories are not necessarily the most important ones for prediction. For example, most of the clothing items that occupy large areas are not important for prediction, since they provide little additional information beyond the presence of persons, a feature that is already included in the model.

To test whether humans and the model differ in the categories they deem important for protest classification, we conduct an additional validation exercise. For this purpose, 1,000 protest images -- 100 for each country -- are randomly selected, and a human coder was asked to name up to three objects that she considers to be most important in recognizing each image as a protest image. Then, the coder was shown the same image with the LVIS segments highlighted and numbered (without labels). The coder then had to identify which objects she selected in the first step correspond to the segments in the second step. Out of 2,210 objects identified in the first step, the five most important ones were people (776), flag (430), poster (216), signboard (195), and banner (118), closely matching categories that our two-level method identifies. An additional test assesses whether all human-coded objects match the segments the machine identifies by comparing the names given by the coder with the ones assigned by the segmenter. When we match strictly based on identical names, 68\% of the human-coded objects are correctly identified by the segmenter. When matching based on a dictionary accounting for membership in the same object categories (i.e., a ``child'' is a ``person''), the successful matching rate increases to 74\%. Appendix G provides further details on this manual validation. Overall, this exercise shows that both human coders and the segmenter largely use the same objects to classify protest images.

\subsection{Country-Specific Importance of Features}

Since protesters employ different tactics across different countries and episodes, the predictive importance of certain features may differ across the countries. For this reason, we assess differences across individual countries, comparing the relevance of a particular image feature in one country in relation to its importance in the whole sample. In the following, we compute this country-specific importance of a feature as its deviation from the average importance of the feature. We again use the best performing model. We are specifically interested in the cases that have a particularly large deviation (positive or negative) for a feature, some of which we discuss in more detail below. 

\begin{figure}[hb!]
    \centering
    \begin{subfigure}{\textwidth}
        \caption{Posters}
        \begin{subfigure}[t]{0.46\textwidth}
        \centering
        \vskip 0pt
        \includegraphics[width=\textwidth]{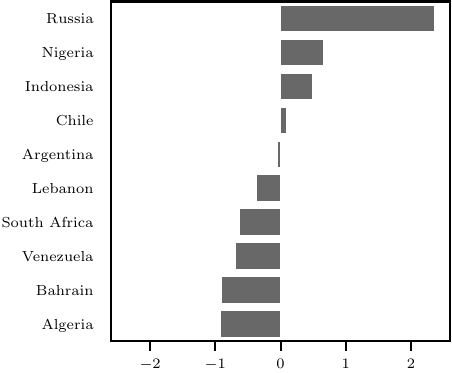}
        \end{subfigure}
        \begin{subfigure}[t]{0.46\textwidth}
            \centering
            \vskip 0pt
            \includegraphics[height=53mm]{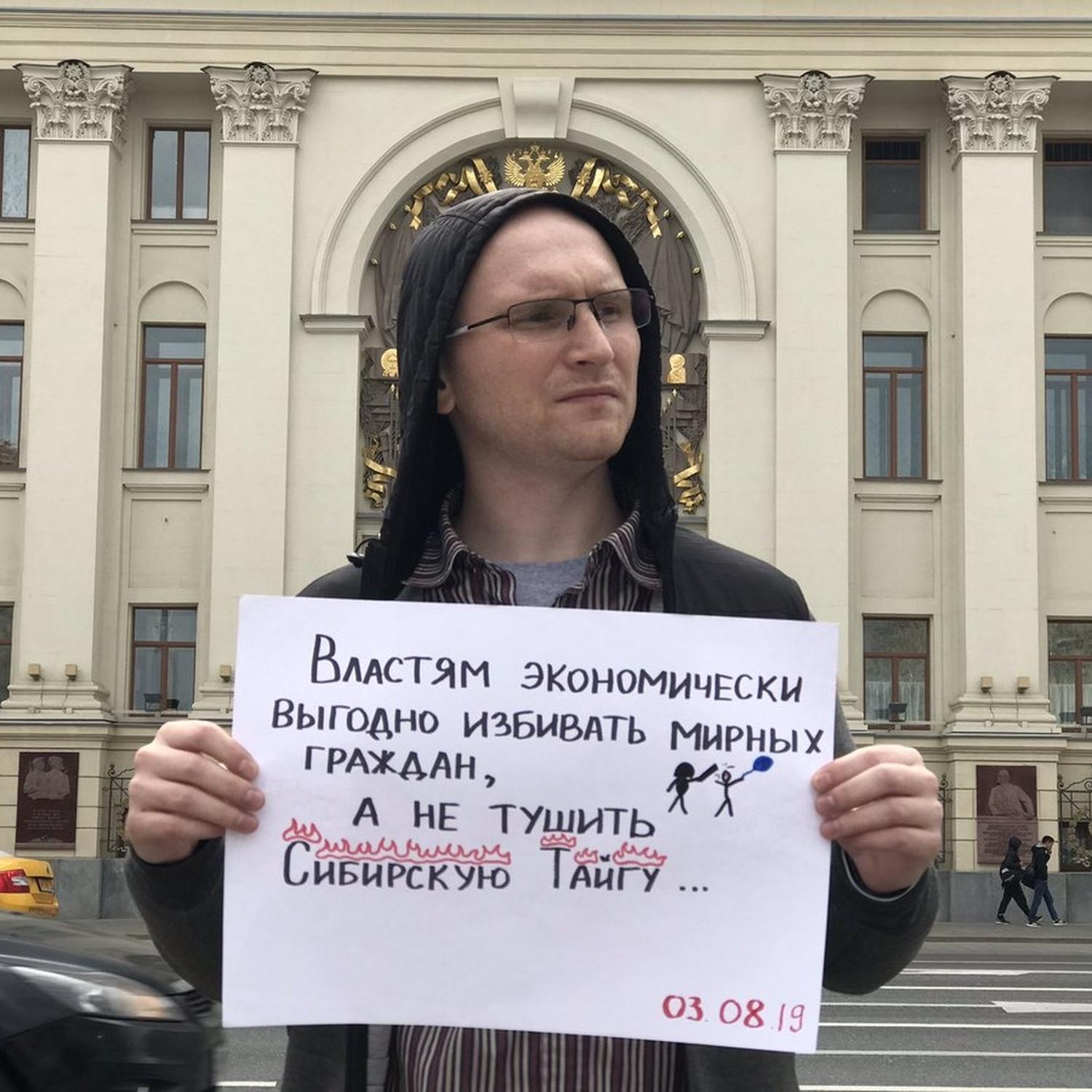}
        \end{subfigure}
    \end{subfigure}

    \begin{subfigure}{\textwidth}
        \caption{Cars}
        \begin{subfigure}[t]{0.46\textwidth}
            \centering
            \vskip 0pt
            \includegraphics[width=\textwidth]{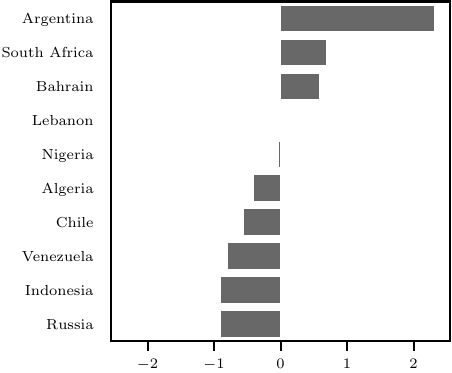}
        \end{subfigure}
        \begin{subfigure}[t]{0.46\textwidth}
            \centering
            \vskip 0pt
            \includegraphics[height=53mm]{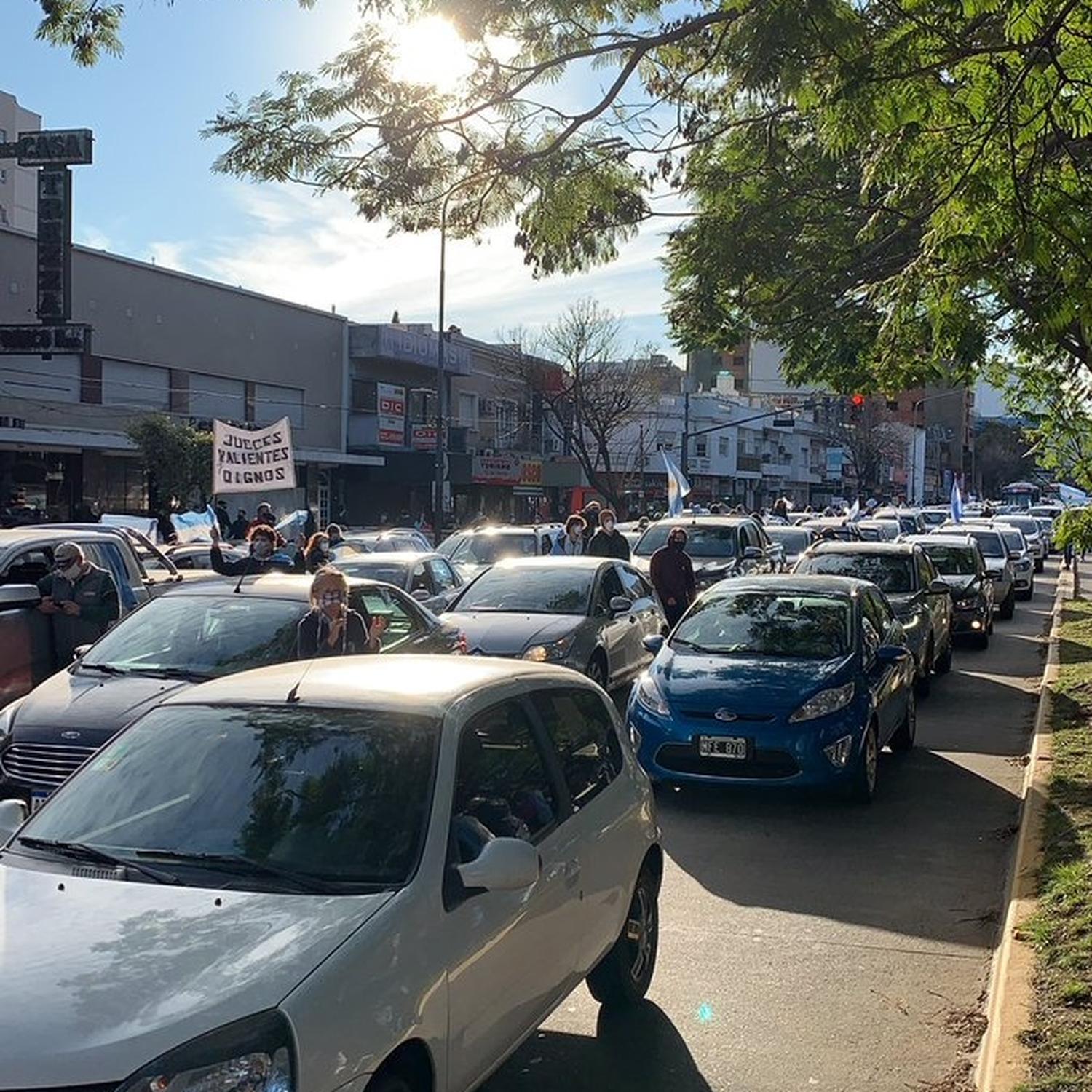}
        \end{subfigure}
    \end{subfigure}
    
    \begin{subfigure}{\textwidth}
        \caption{Candles}
        \begin{subfigure}[t]{0.46\textwidth}
            \centering
            \vskip 0pt
            \includegraphics[width=\textwidth]{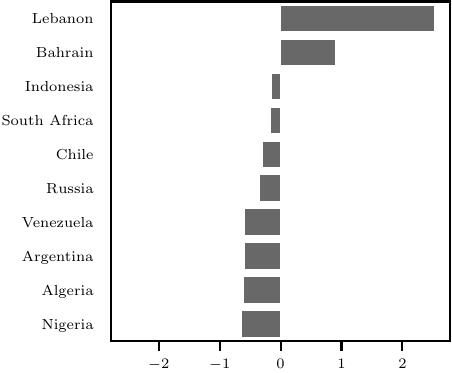}
        \end{subfigure}
        \begin{subfigure}[t]{0.46\textwidth}
            \centering
            \vskip 0pt
            \includegraphics[height=53mm]{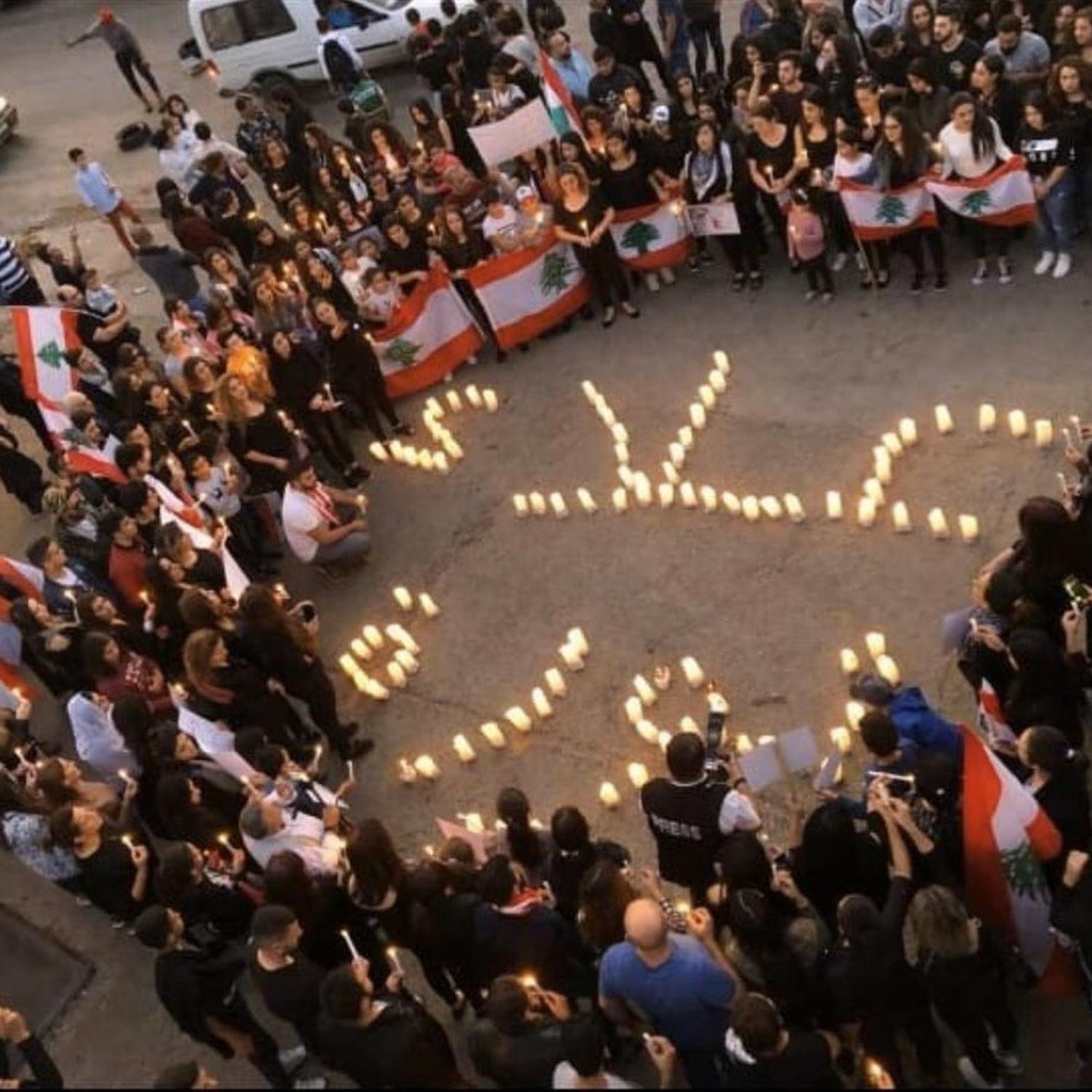}
        \end{subfigure}
    \end{subfigure}
    \caption{Differences in importance of posters, cars and candles across different protest episodes. The left column presents the differences in importance of objects. Positive values denote higher importance in relation to the whole sample. The right column shows examples of protest images: the use of posters in Russia, protest including cars from Argentina, and the use of candles during protest in Lebanon.} 
    \label{fig:features-differences}
\end{figure}

Protesters frequently hold signs stating demands, but the prevalence of signs at protests varies across countries. Signs demanding free elections were an integral part of the protest surge in Moscow that preceded the highly contested 2019 Moscow City Duma elections \citep{roth2019thousands}. The images from these protests confirm that the Russian protests indeed stand out as regards the use of posters in comparison to other protest episodes. Figure \ref{fig:features-differences}, Panel (a) shows the relative importance of signs in Russia and other countries and a sample image of a protester displaying a poster during one contentious episode.

Cars present another feature that characterizes protests in a few countries, but is relatively unimportant for most events in our sample. Argentina experienced a surge of protest in 2020 that mainly targeted governmental responses to the COVID-19 pandemic but also expressed discontent with peoples' economic hardship. As protests took place under governmental quarantine measures requiring Argentinians to socially distance, many protesters turned up in cars during some of these events \citep{bbc2020covid}. Figure \ref{fig:features-differences}, Panel (b) shows a sample image from these protests. In South Africa, on the other hand, cars -- while a significant feature (see Figure \ref{fig:features-differences}, Panel (b)) -- were important for very different reasons. The country experienced a wave of violent unrest during the ``July 2021 riots'' over former president Zuma's imprisonment. During some of these events, rioters looted and burned trucks and cars \citep{guardian2021south}, making these vehicles a crucial feature for the images in our sample. 

As protests often arise as the result of killings or to commemorate deaths, candles are a common feature of protest images. In Lebanon, candles were especially prominent. The ``17 October Revolution'' in Lebanon describes a wave of nationwide anti-government protests in 2019 that led to the resignation of the government. The protests united citizens and groups across the political and religious spectrum demanding a change of the whole political system. Candles played an important role in this context (see Figure \ref{fig:features-differences}, Panel (c)) for two reasons. The shooting of Alaa Abou Fakhr, a peaceful protester, by security forces was followed by public mourning and vigils hold by other protesters who lit candles to commemorate his death \citep{azhari2019}. In addition, lighting candles was an important symbol used by women's marches that were at the core of the movement \citep{mcculloch2019as}. Figure \ref{fig:features-differences}, Panel (c) shows a sample image of an event in Lebanon where candles are central to the scene.

Finally, since tweets contain timestamps, temporal analysis of visual protest features is possible. Figure A4 in Appendix F shows how the prevalence of the three most common segments per country changes before, during, and after a country's protest period. This analysis is meant simply as a proof of concept since this type of analysis is not possible with pixel-based methods' approaches to interpretability. Future research using this method may prefer other methods of selecting segments, such as the ten most common or the five most important.

These cases show that protest can look very different across countries. It shows that the importance of features for some episodes of contention can be highly context-dependent, such as cars during COVID-19 lockdowns or the use of candles to commemorate protesters' death. The examples we discuss also show that some features that are rather widespread nevertheless are of particular importance in other cases as the discussion of posters in Russia has highlighted. Overall, our feature-based analysis considers differences between protest happening in different places and different times, and as a consequence also captures untypical instances of contention and helps to understand why they stand out.

\section{Conclusion}
\label{sec:conclusion}

Images have become increasingly important as a data source for political scientists. Not only has the digital age proliferated their massive spread across the globe, but recent progress in computer vision has also advanced automated analysis of visual material. Existing methods can achieve high accuracy when classifying image content, but their opacity fails to explain \textit{why} classification is made in a certain way. Transparency, however, is important to understand the classification of complex image scenes or assess differences in the same research subject across different cases.

This paper presents a new two-level procedure that improves transparency and interpretability of image classification. The first step extracts objects from an image and creates an abstract representation of the image based on these segments. This procedure improves on bag of visual words (BoVW) methods because it identifies human understandable items in an image; in contrast, BoVW identifies areas of high pixel contrast and represents these areas as vectors based on changes in pixel intensity \citep{torres2023bovw}. While BoVW is a significant improvement over convolutional neural network and transformer classifiers, its reliance on pixel clusters means it is not as interpretable as this paper's method. In the second step, these feature vectors are then used in standard machine learning classifiers to predict the image content. This method and its advantages over previous approaches is demonstrated in an application to protest image analysis. Every day, people take to the streets to protest and images of these protests are shared widely on social media. Existing methods are able to classify accurately this material as protest images, but they make it difficult to understand what particular content in an image depicting a contentious scene leads the algorithm to predict protest. Our method addresses this challenge. 

Our two-level approach achieves a slightly lower predictive performance than state-of-the-art image classification methods, but it has three advantages. First, simply knowing which objects are in a protest image is an advance in transparency over conventional computer vision models. Existing approaches emphasizes pixels, but pixels do not often map to human understandable concepts. In addition, the abstract descriptions of the images -- i.e., the segments they contain -- can be shared as part of the replication material, which often does not include the images themselves for privacy and copyright reasons. There is no equivalent for pixel-based vision models. The closest, sharing the image embedding, places an image in multidimensional space but does not reveal anything about what the image contains. Second, the researcher can study which objects and associated features typically lead to an image being classified in a particular way. For protests, people and items such as a banners and signs distinguish protest from non-protest images. Third, this method can be used to study protest features specific to particular events. We show how our method can be used to identify objects that are predictive of protest only in particular countries, which enables comparative research on protest tactics. Hence, we demonstrate that the increasing focus on interpretable AI \citep{guidotti2019survey} is not only an aim in itself, but can lead to new and improved research that conventional methods cannot be used for.

Several extensions are possible. First, all our feature generation methods use \textit{atomic} objects, i.e.~some measure of the object independent of other types of objects. Features could also be generated which represent dyadic relationships, for example, how far an object of a certain type is located away from an object of another type. However, the combinatorial explosion of the feature space makes this difficult. Second, researchers can rely on other image segmentation methods, including those provided by commercial actors such as Amazon's Rekognition. We did not do this on purpose, and chose to rely on open-source products that are available to other researchers (at no or low cost), and which allow for the replication of our results. If researchers choose to prioritize speed over transparency and replication, commercial tools can be an option at the first stage. Third, rather than using a generic vocabulary, objects and feature generations could be used that are based on theoretical priors, where one constrains the object (or feature) space because of existing theoretical or qualitative knowledge. In contrast to our inductive approach that requires no prior knowledge, the use of domain-specific segment categories can likely increase performance for particular applications. At the same time, however, constraining will result in a model that is custom-tailored to specific phenomena the researcher is interested in and hence no longer as generally applicable as our method. Fourth, there is ongoing work on open vocabulary segmentation to let the user specify their own vocabulary rather than using a pre-specified one \citep[e.g.][]{zhou2022detecting}. For example, LVIS (or any other vocabulary) does not identify police officers or fire, common components of protest imagery. Fifth, and most importantly, the application of this paper's method is not limited to the study of political protest. In fact, any social science image classification task that relies on complex scenes, such as politicians' campaign imagery \citep{xi2020understanding}, may be performed with this two-level approach.

At a practical level, readers may wonder about the availability of images for research purposes in general. As a result of Elon Musk's purchase of Twitter, the ability to collect large numbers of images from there on an academic's budget has been severely curtailed. Appendix H explains these new restrictions in more detail as well as three alternatives: using images from already downloaded tweets, scraping, or working within the European Union's Digital Services Act. Though research with Twitter is no longer as easy as it used to be, the future of research with images remains as promising as ever.  

\section*{Funding}
This project was funded by a research award from the Dr.~K.~H.~Eberle Foundation. 

\section*{Acknowledgements}
The authors gratefully acknowledge comments from Andreu Casas, Nora Webb-Williams, and Han Zhang, as well as Michelle Torres at the 2023 Annual Meeting of the American Political Science Association.

\section*{Data Availability Statement}
Replication code, model weights and data for this article will be published on Dataverse. To enable users to apply the method to their own images, we provide an interactive demonstration application with reduced functionality (\url{https://huggingface.co/spaces/ciass/protest-segments}) and an API (\url{https://huggingface.co/spaces/ciass/protest-segments?view=api}) via Hugging Face.

\printbibliography

\end{document}


\maketitle

\onehalfspacing

\beginappendix

\section{Protest Image Dataset}
\label{app:coding}

The Protest Image Dataset is a new data collection project that includes images from social media with an emphasis on political protests. The dataset contains, besides the images themselves, variables on location, time, and a hand-annotated protest variable. This section describes the general conventions guiding the image collection and image annotation.

\subsection{Image Collection}
\label{app:coding-selection}

\paragraph{Selection of Countries}

We used the Armed Conflict, Location, and Event Dataset (ACLED) to analyze all protests since January 1, 2014 to the present \citep{raleigh2010introducing}. We then identified the twenty country-years with the most protest events for each of ACLED's 16 regions, resulting in 313 candidate country-years\footnote{ACLED did not start covering North America until Mexico in 2018; its coverage guide lists Mexico as Central America, but in the dataset the country is coded as North America. The United States was added in 2020, Bermuda, Canada, and Saint Pierre and Miquelon in 2021. Every candidate country-year for North America was therefore included at this stage of the selection.}. Next, the logistic regression model from \citet{steinert-threlkeld2022how} was used to identify the 46 countries (171 country-years) with enough people and income to produce enough protest images from Twitter. These 46 were narrowed to 14 based on their Polity IV score and region, with a goal of generating broad coverage of regime types and parts of the world. With 14 countries we could ensure that we annotate a sufficient number of images per country despite the restriction due to the labor-intensive annotation of images.

\paragraph{Selection of Posts}

Because of its widespread use throughout the world \citep{huang2019large}, we used the social media platform Twitter to obtain protest images posted by observers on the ground. In order to assign a tweet to a specific country, we required that the tweet was geolocated within this country. Though it is possible that users who geolocate their tweets are not a representative sample of their country’s population \citep{malik2021population}, work comparing Twitter users who share protest images to those who share non-protest images finds no differences between those two groups \citep{steinert-threlkeld2022how}. As the storage requirements would render it impossible for us to collect within a whole year all geolocated tweets, particularly in the large countries, we decided to narrow down the tweets for each country to a specific date range. Thus, we continued to analyze the 14 countries’ number of protests per month, and chose a date range that includes both the rise and the fall of the protests. For most countries, we specified the start date on the first of the rising month and the end date on the last of the falling month. We made an exception for countries where we expected a particularly large number of tweets; we specified their start and end dates also within the courses of these months. For all countries, we ensured that within these date ranges, in addition to tweets posted during high numbers of protests, we also included tweets posted seven days before and after the protest period. After selecting these periods, we extracted tweets from the relevant country-days from a corpus of tweets downloaded from Twitter’s POST statuses/filter endpoint. This extraction resulted in just over 135 million tweets which were then used to find protest images.

\paragraph{Selection of Images}

Twitter allows a tweet to have multiple media; it allows up to four photos, one animated GIF or one video. In our selection of images, we included all media that Twitter categorized as a photo, but no media was categorized as a GIF or video. We then downloaded these images and saved them together with their tweet identifier, tweet date and media identifier. Despite our previous selection of tweets, in some countries we collected far too many images to store them in the space available to us, not to mention annotate them in the next step. Therefore we decided to introduce a limit of images per country at 100,000. This limit affected 11 countries, where to stay below the limit we randomly sorted the images and then downloaded them until we had 100,000. For example, in Japan, the country where we collected the most images, 5,546,059 images were sampled to 100,000. In contrast, the Kazakhstan tweets contained only 52,825 images, so we kept all of them.

\begin{table}[t]
\centering
\begin{tabular}{llllr}
\toprule
 & Region & Country & Date range & Images \\
\midrule
1 & Northern Africa & Algeria & 2019-02-01 – 2020-03-01 & 100,000\\
2 & Middle East & Lebanon & 2019-10-01 – 2020-01-15 & 42,203\\
3 & Middle East & Bahrain & 2016-01-01 – 2017-12-31 & 100,000\\
4 & South America & Argentina & 2020-05-01 – 2020-09-30 & 100,000\\
5 & South America & Chile & 2019-10-01 – 2019-12-31 & 100,000\\
6 & South America & Venezuela & 2019-01-01 – 2019-11-30 & 100,000\\
7 & Eastern Africa & Ethiopia & 2015-11-01 – 2016-12-31 & 5,867\\
8 & Southern Africa & South Africa & 2021-01-01 – 2021-08-31 & 100,000\\
9 & Western Africa & Nigeria & 2018-09-01 – 2019-09-30 & 100,000\\
10 & Caucasus and Central Asia & Kazakhstan & 2019-01-01 – 2020-03-30 & 52,825\\
11 & Europe & Russia & 2019-07-07 – 2019-10-06 & 100,000\\
12 & Southeast Asia & Indonesia & 2019-05-01 – 2019-10-31 & 100,000\\
13 & East Asia & Japan & 2018-02-22 – 2018-06-30 & 100,000\\
14 & Southeast Asia & Philippines & 2017-05-01 – 2017-12-31 & 100,000\\
\bottomrule
\end{tabular}
\caption[Selection of images]{Selection of images}
\label{tab:collection}
\end{table}

\subsection{De-Duplication}
\label{app:coding-deduplication}

We analyze the occurrence of duplicates in our dataset to rule out possible problems: Through many duplicates of the same image, the importances of certain features could be inflated. In addition, if the same image occurs in the training set as well as in the testing set, the classification results would be biased. We generate encodings for the images by propagating them through a convolutional neural network. We use a MobileNet v3 \citep{howard2017mobilenets} pretrained on the ImageNet dataset \citep{deng2009imagenet} and sliced at the last layer. This generates an encoding of 576 features. We compute the cosine similarity between all pairs of images and retrieve duplicates with a similarity equal or larger than 0.99. We then identify 127,769 duplicate images in 48,905 clusters. The duplicate images are dropped from the dataset.

\subsection{Image Annotation}
\label{app:coding-annotation}

We annotate images as to whether they display a political protest, or part of it. We define protest as

\begin{itemize}
    \item\textbf{A publicly visible event or action}: It takes place in a public space and therefore can be observed by the public.
    \item\textbf{An event involving one or more participants that are present on site}: Protest can range from individual statements to mass demonstrations. We exclude instances where a symbol or an item is displayed publicly without the presence of protesters themselves.
    \item\textbf{A political statement or expression}: An objection or a criticism against a political actor or institution. This can be achieved by means of anything not corresponding to the norm and thus attracting public attention; it can be done by verbal statements or speeches, but also with banners or symbols.
\end{itemize}

Images often cover protests only partially; for example, they display a single person or a group of persons participating in the protest. These images are considered “protest” images, if their relation to a protest as defined above can be ascertained. They do not need to display a complete protest event. The coder’s annotation is coded on a four-point scale as

\begin{itemize}
    \item\textbf{Protest (high certainty)}: The coder is certain that the image shows a part of a protest as defined above.
    \item\textbf{Protest (low certainty)}: The coder believes that the image probably shows a part of a protest as defined above.
    \item\textbf{No protest (low certainty)}: The coder believes that the image probably does not show a part of a protest as defined above. statements or speeches, but also with banners or symbols.
    \item\textbf{No protest (high certainty)}: The coder is certain that the image does not show a part of a protest as defined above.
\end{itemize}

We present the coders in the first round with 6,000 images from each country. These images are randomly selected from the previously selected images. In the second, third and fourth round, we select from each country 3,000 images by weighted random sampling. To calculate the weights, we train a model on the already annotated images. This model is based on a vision transformer \citep[ViT,][]{dosovitskiy2020image}; it is retrained after each round of annotations. This model gives us for every not-yet-annotated image a score between 0 and 1, where a low score indicates a likely-non-protest image and a high score a likely-protest image. The images are then grouped by these scores in 20 equal-width bins, and their weights are calculated such that the probability of drawing an image from one bin is the same as from another bin. The aim of this weighted random sampling is to reduce the probability of likely-no-protest images and increase the probability of likely-protest images. The annotation of the images proceeds until the coders' available time is used up. 

\subsection{Analyzing Reliability Across Coders}
\label{app:coding-reliability}

We analyze the degree that coders consistently assigned categorical protest ratings to the images in our dataset. The protest annotations for 141,538 images were done by four coders. For 65,120 images we have annotations from two coders.

Cohen's kappa was computed for four classes (no protest high, no protest low, protest low, protest high), with an inter-rater reliability of 0.68. When we combine high and low confidence ratings to obtain a binary classification, we obtain an inter-rater reliability of 0.81. According to \citet{mchugh2012interrater}, these results indicate moderate and strong reliability, respectively. Since each image is annotated by a random set of coders, we decide to also compute the intraclass correlation correlation coefficient (ICC1). This is equal to a one-way ANOVA fixed effects model. For four classes (no protest high, no protest low, protest low, protest high) this gives us a intraclass correlation of 0.83. For two classes (no protest, protest) this gives us a intraclass correlation of 0.79, which is a good result according to \citet{koo2016guideline}. 

\subsection{Splitting Images into Training and Testing Set}
\label{app:coding-splitting}

Table \ref{tab:data_full} presents the number of images per country in the four annotation categories. These annotated images in the dataset are randomly split into a training and testing set. The training set contains 80\% of the images, whereas the testing set contains 20\%. 

\begin{table}[h!]
\centering
\small
\begin{tabular}{lrrrr}
\toprule
 & \multicolumn{2}{c}{No Protest} & \multicolumn{2}{c}{Protest} \\
 & High & Low & Low & High \\
\midrule
Argentina & 12,864 & 124 & 343 & 1,087 \\
Bahrain & 14,070 & 156 & 111 & 172 \\
Chile & 9,708 & 445 & 1,016 & 2,776 \\
Algeria & 9,756 & 214 & 617 & 2,450 \\
Indonesia & 13,845 & 267 & 215 & 434 \\
Lebanon & 11,082 & 279 & 586 & 2,128 \\
Nigeria & 13,565 & 222 & 177 & 309 \\
Russia & 14,040 & 113 & 135 & 300 \\
Venezuela & 11,006 & 322 & 656 & 1,790 \\
South Africa & 13,815 & 78 & 94 & 171 \\
\midrule
Total & 123,751 & 2,220 & 3,950 & 11,617 \\
\bottomrule
\end{tabular}
\caption[Protest image dataset]{Images in protest images dataset annotated in different protest categories and countries.}
\label{tab:data_full}
\end{table}

\clearpage

\section{Training of Models}
\label{app:training}

For the models we train using our segment-based approach, we choose four different classification methods: logistic regression, simple decision trees, collections of decision trees and gradient-boosted decision trees. We use logistic regression because it is widely used by social scientists, and to provide a benchmark against. We choose the tree-based models because they intuitively allow us to vary the complexity and interpretability of the models. As implementation of the collections of decision trees a random forest \citep{breiman2001random} is used, for gradient-boosted decision trees XGBoost \citep{chen2016xgboost} is used. 

In order to make a comparison with conventional computer vision methods, we also have to make a selection of these methods. We decide to train a convolutional neural network (CNN) by ourselves. We decide to train a ResNet50 because we want to keep the training times and hardware requirements lower compared to, for instance, a ResNet101 or Resnet152 \citep{he2016deep}. It also allows a direct comparison to the same architecture but trained on a different dataset by \citet{won2017protest}. In addition, we select a vision transformer (ViT) because they have shown to outperform CNNs on many computer vision tasks while requiring less computational resources \citep{dosovitskiy2020image}. These vision transformers are available as base, large and huge-sized variants. We make sure to use a base-sized variant of the ViT to make the comparison to the Resnet50 as fair as possible. Our ViT model refers to a base-sized variant with a patch resolution of 16x16 and a fine-tuning resolution of 384x384.

The first step in the training of each segment model is to select the hyperparameters. For this purpose, a 5-fold cross-validation is performed for the complete grid of hyperparameters. For the logistic regression, different regularization strengths are tried, with up to 10 improving the accuracy. For the simple decision trees, the maximum depth is varied from 1 to 16. From a depth of 8 to 16, most classifiers improve only minimally, or even deteriorate. For the random forests, the number of trees, the number of maximum features, the maximum depth and the minimum number of samples in a leaf are varied. The number of trees is varied from 1 to 1,000, with more trees leading to no obvious improvement. For the gradient-boosted trees, a large number of hyperparameters is varied, the maximum depth, the number of boosting rounds, learning rate, and minimum loss reduction. If we disable boosting (number of boosting rounds 0), the maximum F1 score is achieved with a maximum depth of 8. The score deteriorates if the maximum depth is above or below 8. In order to look at the effect of the number of boosting rounds, we fix the maximum depth at 8. The F1 scores improve with more boosting rounds, until 10,000 boosting rounds. 

For the training of the ResNet50 and ViT, we use pre-trained weights on the ImageNet dataset. This way, the model knows from the beginning certain features that are independent of our protest images, such as corners, edges, shapes, etc. We never use the trained weights of the ResNet50 by \citet{won2017protest}, also not as pre-trained weights for our self-trained ResNet50. During the training of the models, however, these pre-trained weights could be completely changed, as no layers are frozen and the gradients for all weights in all layers are calculated and changed. For the sake of readability, we have decided to use the term training. But by the definition of finetuning, this ``training" procedure could also be referred to as ``finetuning" procedure. We decide to use a cross entropy loss with a stochastic gradient descent optimizer with momentum. For hyperparameter tuning, the training data is additionally split into a training (80\%) and validation set (20\%). This is not the same as the 5-fold cross validation for the segments models, but fulfills a similar purpose with significantly less computational effort. We follow best practice for setting most of the hyperparameters. But we optimize the values for the learning rate and momentum with the help of hyperparameter tuning. It is found that a learning rate of 1e-03 is best for the ResNet50, while it is significantly lower for the ViT at 1e-05. The optimal momentum is found to be 0.99 for the ResNet50 and 0.99 for the ViT. After the optimal hyperparameters for the models are found, they are retrained on the entire training data for 100 epochs.

All models are trained on a server node with 8 Intel Xeon @ 2.50 GHz cores, 128 GB memory as well as a NVIDIA graphics card, Quadro RTX 6000 with 24 GB memory. From the segment models, the logistic regression and gradient-boosted trees need the longest training time -- but not more than 12 minutes. The final model of the ResNet50 is trained in 15 hours, while the ViT is trained in 30 hours. To infer whether the images in our dataset are protest images, the gradient-boosted tree needs 2 minutes (0.0007 seconds/image). The inference with the ResNet50 needs 6 minutes (0.0025 seconds/image) and the ViT 30 minutes (0.0129 seconds/image).

\clearpage
\section{Main Results}
\label{app:mainresults}

We provide the full results for the full combination of different design choices in Table \ref{tab:models}. Results for the conventional image classification methods are provided at the bottom of the table. 

\begin{table}[h!]
\centering
\small
\begin{tabular}{lrrrrrr}
\toprule
 & \multicolumn{3}{c}{Training} & \multicolumn{3}{c}{Testing} \\
 & Precision & Recall & F1 & Precision & Recall & F1 \\
\midrule
Segments (COCO, bin, logistic) & 0.6495 & 0.3118 & 0.4213 & 0.6249 & 0.3013 & 0.4066 \\
Segments (COCO, count, logistic) & 0.6527 & 0.4621 & 0.5411 & 0.6589 & 0.4729 & 0.5506 \\
Segments (COCO, area max, logistic) & 0.1691 & 0.0028 & 0.0055 & 0.1667 & 0.0026 & 0.0051 \\
Segments (COCO, area sum, logistic) & 0.2289 & 0.0067 & 0.0131 & 0.2135 & 0.0061 & 0.0119 \\
\midrule
Segments (COCO, bin, tree) & 0.7672 & 0.3713 & 0.5004 & 0.5735 & 0.2856 & 0.3813 \\
Segments (COCO, count, tree) & 0.5225 & 0.7304 & 0.6092 & 0.5187 & 0.7180 & 0.6023 \\
Segments (COCO, area max, tree) & 0.7629 & 0.5931 & 0.6674 & 0.5163 & 0.4019 & 0.4520 \\
Segments (COCO, area sum, tree) & 0.7642 & 0.5741 & 0.6557 & 0.5376 & 0.4022 & 0.4601 \\
\midrule
Segments (COCO, bin, forest) & 0.8011 & 0.2658 & 0.3991 & 0.6762 & 0.2120 & 0.3228 \\
Segments (COCO, count, forest) & 0.8082 & 0.5625 & 0.6633 & 0.6984 & 0.4767 & 0.5666 \\
Segments (COCO, area max, forest) & 0.6486 & 0.3249 & 0.4329 & 0.5527 & 0.2830 & 0.3743 \\
Segments (COCO, area sum, forest) & 0.8400 & 0.3608 & 0.5047 & 0.6808 & 0.2679 & 0.3845 \\
\midrule
Segments (COCO, bin, boosted tree) & 0.7362 & 0.4567 & 0.5637 & 0.6237 & 0.3701 & 0.4645 \\
Segments (COCO, count, boosted tree) & 0.7091 & 0.5860 & 0.6417 & 0.6836 & 0.5699 & 0.6216 \\
Segments (COCO, area max, boosted tree) & 0.9156 & 0.7140 & 0.8023 & 0.6374 & 0.4330 & 0.5157 \\
Segments (COCO, area sum, boosted tree) & 0.8129 & 0.5734 & 0.6724 & 0.6486 & 0.4305 & 0.5175 \\
\midrule
Segments (LVIS, bin, logistic) & 0.7509 & 0.5831 & 0.6565 & 0.7389 & 0.5818 & 0.6510 \\
Segments (LVIS, count, logistic) & 0.7254 & 0.5138 & 0.6016 & 0.7048 & 0.5130 & 0.5938 \\
Segments (LVIS, area max, logistic) & 0.4443 & 0.0480 & 0.0867 & 0.4259 & 0.0443 & 0.0803 \\
Segments (LVIS, area sum, logistic) & 0.5374 & 0.1137 & 0.1877 & 0.5521 & 0.1124 & 0.1868 \\
\midrule
Segments (LVIS, bin, tree) & 0.8976 & 0.8005 & 0.8463 & 0.5942 & 0.5429 & 0.5674 \\
Segments (LVIS, count, tree) & 0.8831 & 0.7778 & 0.8271 & 0.6476 & 0.5667 & 0.6044 \\
Segments (LVIS, area max, tree) & 0.7650 & 0.5149 & 0.6155 & 0.7154 & 0.4626 & 0.5618 \\
Segments (LVIS, area sum, tree) & 0.9219 & 0.8364 & 0.8771 & 0.6097 & 0.5596 & 0.5836 \\
\midrule
Segments (LVIS, bin, forest) & 0.9568 & 0.5087 & 0.6642 & 0.8415 & 0.3736 & 0.5175 \\
Segments (LVIS, count, forest) & 0.9592 & 0.5719 & 0.7165 & 0.8256 & 0.4333 & 0.5684 \\
Segments (LVIS, area max, forest) & 0.9431 & 0.5416 & 0.6881 & 0.7817 & 0.3762 & 0.5079 \\
Segments (LVIS, area sum, forest) & 0.9761 & 0.5455 & 0.6999 & 0.8457 & 0.3784 & 0.5229 \\
\midrule
Segments (LVIS, bin, boosted tree) & 0.9944 & 0.9753 & 0.9848 & 0.7594 & 0.6315 & 0.6896 \\
Segments (LVIS, count, boosted tree) & 0.9982 & 0.9813 & 0.9897 & 0.7834 & 0.6624 & 0.7178 \\
Segments (LVIS, area max, boosted tree) & 1.0000 & 1.0000 & 1.0000 & 0.7805 & 0.6569 & 0.7134 \\
\fontseries{b}\selectfont Segments (LVIS, area sum, boosted tree) & \fontseries{b}\selectfont 1.0000 & \fontseries{b}\selectfont 1.0000 & \fontseries{b}\selectfont 1.0000 & \fontseries{b}\selectfont 0.7821 & \fontseries{b}\selectfont 0.6675 & \fontseries{b}\selectfont 0.7203 \\
\midrule
ResNet50 (Won et al., 2017) & 0.5834 & 0.4698 & 0.5205 & 0.5787 & 0.4584 & 0.5116 \\
ResNet50 (self-trained) & 0.8508 & 0.8192 & 0.8347 & 0.7657 & 0.7327 & 0.7489 \\
\fontseries{b}\selectfont ViT (self-trained) & \fontseries{b}\selectfont 0.9199 & \fontseries{b}\selectfont 0.8939 & \fontseries{b}\selectfont 0.9067 & \fontseries{b}\selectfont 0.8400 & \fontseries{b}\selectfont 0.8060 & \fontseries{b}\selectfont 0.8226 \\
\bottomrule
\end{tabular}
\caption[Evaluation of different methods]{Evaluation of different methods. ``Self-trained'' means trained on the images collected for this project.}
\label{tab:models}
\end{table}

To compare the performance between countries, we present results for the best conventional method (vision transformer model, ViT) and the best of our two-level classifiers (LVIS vocabulary, area sum features and a boosted tree classifier). Table \ref{tab:model_country} presents the results for the images in the 10 countries of our dataset. 

\begin{table}[hb!]
\centering
\small
\begin{tabular}{llrrrrrr}
\toprule
 &  & \multicolumn{3}{c}{Training} & \multicolumn{3}{c}{Testing} \\
 &  & Precision & Recall & F1 & Precision & Recall & F1 \\
\midrule
\multirow[c]{10}{*}{Segments} & Argentina & 1.0000 & 1.0000 & 1.0000 & 0.7500 & 0.6167 & 0.6769 \\
 & Bahrain & 1.0000 & 1.0000 & 1.0000 & 0.3803 & 0.4821 & 0.4252 \\
 & Chile & 1.0000 & 1.0000 & 1.0000 & 0.8560 & 0.5884 & 0.6974 \\
 & Algeria & 1.0000 & 1.0000 & 1.0000 & 0.8933 & 0.7651 & 0.8243 \\
 & Indonesia & 1.0000 & 1.0000 & 1.0000 & 0.5490 & 0.6462 & 0.5936 \\
 & Lebanon & 1.0000 & 1.0000 & 1.0000 & 0.8646 & 0.7053 & 0.7769 \\
 & Nigeria & 1.0000 & 1.0000 & 1.0000 & 0.5463 & 0.6082 & 0.5756 \\
 & Russia & 1.0000 & 1.0000 & 1.0000 & 0.5488 & 0.5172 & 0.5325 \\
 & Venezuela & 1.0000 & 1.0000 & 1.0000 & 0.7804 & 0.7342 & 0.7566 \\
 & South Africa & 1.0000 & 1.0000 & 1.0000 & 0.5000 & 0.5472 & 0.5225 \\
\midrule
\multirow[c]{10}{*}{ViT} & Argentina & 0.9019 & 0.8924 & 0.8971 & 0.8284 & 0.7735 & 0.8000 \\
 & Bahrain & 0.8423 & 0.8238 & 0.8330 & 0.6667 & 0.6786 & 0.6726 \\
 & Chile & 0.9196 & 0.8708 & 0.8945 & 0.8569 & 0.7586 & 0.8048 \\
 & Algeria & 0.9552 & 0.9560 & 0.9556 & 0.9103 & 0.9103 & 0.9103 \\
 & Indonesia & 0.8755 & 0.7996 & 0.8359 & 0.7731 & 0.7077 & 0.7390 \\
 & Lebanon & 0.9187 & 0.9056 & 0.9121 & 0.8396 & 0.8287 & 0.8341 \\
 & Nigeria & 0.8834 & 0.7789 & 0.8279 & 0.7065 & 0.6701 & 0.6878 \\
 & Russia & 0.9217 & 0.8793 & 0.9000 & 0.7674 & 0.7586 & 0.7630 \\
 & Venezuela & 0.9225 & 0.9121 & 0.9173 & 0.8343 & 0.8446 & 0.8394 \\
 & South Africa & 0.8274 & 0.7689 & 0.7971 & 0.6000 & 0.5660 & 0.5825 \\
\bottomrule
\end{tabular}
\caption[Evaluation of different methods per country]{Evaluation of different methods per country. The Vision Transformer (ViT) is the best conventional method, whereas Segments is the best of our two-level classifiers with the LVIS vocabulary, area sum features and a boosted tree classifier.}
\label{tab:model_country}
\end{table}

\clearpage

\section{Analysis of Clustered Images}
\label{app:model_cluster}

We analyze the performance of our classifier on subcategories of protest images. To identify these subcategories, we use an unsupervised approach that clusters the images and thus assigns them to unlabeled categories (see \citet{zhang2022image}).

In order to do this, we extract an embedding for each image in our dataset. This embedding is generated by our self-trained vision transformer (ViT) in the last linear layer, and is 768 features long. We then cluster the embeddings using the Euclidean distance and the KMeans algorithm. To determine the number of clusters, the number of clusters is raised as long as the coherence of each cluster is given. This is done according to the procedure proposed by \citet{zhang2022image} by always selecting 20 random images from each cluster, determining a topic for that cluster, and checking if at least 50\% of the images in that cluster match the topic. This procedure leads to 30 clusters.

Then, we evaluate the accuracy separately on each cluster. As classifier we use our best two-level classifier (LVIS vocabulary, area sum features and a boosted tree classifier). We do not analyze the accuracy on the training images, as these images are all correctly classified and are therefore also correctly classified in the individual clusters. Instead, the accuracy on the test images is analyzed based on the true negatives, false negatives, true positives, false positives, precision score, recall score and F1 score in the individual clusters.

Table \ref{tab:model_cluster} shows the accuracy of the classifier in the clusters that contain at least 20 protest images from the test set. In the cluster of protest images with flags, the precision and recall score are close to each other, which indicates that the classifier is balanced to make errors in classifying either as a protest image and a non-protest image. In the other clusters, however, the precision is higher than the recall. This indicates that in these clusters the classifier makes more errors in classifying non-protest images as protest images than protest images as non-protest images. At the same time, this shows that the accuracy differs between the clusters. The differences in the precision and recall scores considered above are also reflected in the F1 scores. By comparing the F1 scores of the clusters, we see that an F1 score of 0.4156 is achieved for the clusters with African gatherings, with fire smoke of 0.5039, state police of 0.5221 and gatherings of 0.6222. This means that for these clusters it is below the F1 score of 0.7203, which the classifier achieves on all test images. In contrast, it achieves a higher accuracy for the clusters with large mass protests with an F1 score of 0.7495, protests with signboards with 0.8137 and protests with flags with 0.9370. The less accurate clusters can possibly be explained by the fact that object categories are missing for them in the LVIS vocabulary. For example, the vocabulary contains no categories related to fire, smoke and policemen. The difficulties with gatherings could be explained by the fact that it is difficult to distinguish whether it is a protest image or a non-protest image based on the number of people. But especially if there are large masses on the images, the classifier has a good performance, also if objects and flags can be seen on the protest images.

\begin{table}[h!]
\centering
\small
\begin{tabular}{llrrrrrrr}
\toprule
 & & TN & FN & TP & FP & Precision & Recall & F1 \\
\midrule
9 & Edited images & 1,268 & 31 & 17 & 17 & 0.5000 & 0.3542 & 0.4146 \\
10 & Streets & 1,161 & 55 & 41 & 39 & 0.5125 & 0.4271 & 0.4659 \\
14 & Protest with flags & 14 & 58 & 714 & 38 & 0.9495 & 0.9249 & 0.9370 \\
16 & Gathering & 414 & 137 & 210 & 118 & 0.6402 & 0.6052 & 0.6222 \\
17 & Fire smoke & 242 & 101 & 64 & 25 & 0.7191 & 0.3879 & 0.5039 \\
18 & African gatherings & 504 & 19 & 16 & 26 & 0.3810 & 0.4571 & 0.4156 \\
22 & Large mass protests & 108 & 167 & 365 & 77 & 0.8258 & 0.6861 & 0.7495 \\
23 & State police & 288 & 74 & 59 & 34 & 0.6344 & 0.4436 & 0.5221 \\
24 & Protest with signboards & 120 & 174 & 450 & 32 & 0.9336 & 0.7212 & 0.8137 \\
27 & Flags & 287 & 88 & 94 & 46 & 0.6714 & 0.5165 & 0.5839 \\
30 & Random images with text & 1,884 & 67 & 31 & 44 & 0.4133 & 0.3163 & 0.3584 \\
\bottomrule
\end{tabular}
\caption[Evaluation of classifier per cluster]{Evaluation of best two-level classifier for clusters that contain at least 20 protest images from the test set. Evaluation metrics are true negatives (TN), false negatives (FN), true positives (TP), false positives (FP), precision score, recall score and F1 score.}
\label{tab:model_cluster}
\end{table}

In addition to the clusters shown in Table \ref{tab:model_cluster}, there are also clusters that contain less than 20 protest images. These include, for example, a cluster with football matches in which the classifier correctly classifies 406 non-protest images, misclassifies 1 protest image as non-protest image and incorrectly classifies 7 non-protest images. In this case, the F1 score is not defined because there are no true positive cases, which is why precision and recall are zero and the F1 score is not defined. Also in the cluster of concert images 450 non-protest images are classified correctly, 10 non-protest images are classified incorrectly, 7 protest images are classified incorrectly, 1 protest image is classified correctly. This leads to a precision score of 0.0909, a recall score of 0.1250 and an F1 score of 0.1053. These low scores can also be explained by the small number of protest images in this cluster.

To get a visual impression of the clusters, we select sample images from the clusters. To get a representative impression of the clusters, the images are selected according to the centrality in the cluster. For this purpose, the distances of the images to the cluster centroid are calculated in each cluster. Images whose distances are lower are more central in the cluster, whereas images whose distances are higher are further outside the cluster.

\begin{figure}[h!]
    \centering
    \includegraphics[width=0.85\textwidth]{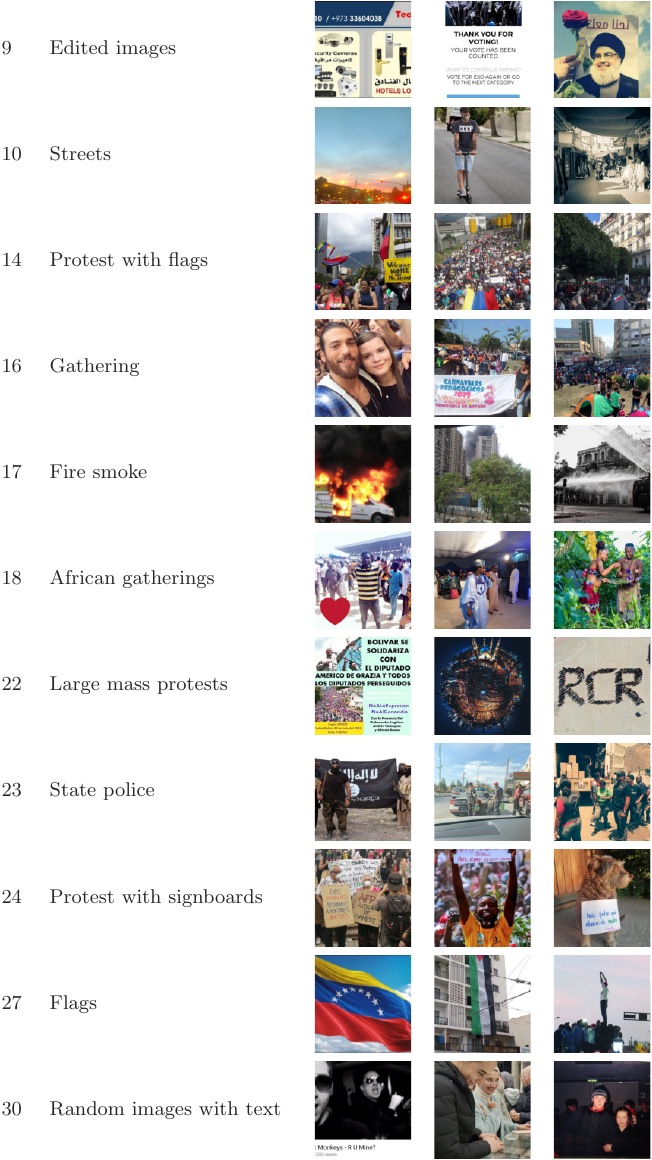}
    \caption[Sample images per cluster]{Sample images for clusters than contain at least 20 protest images from the test set. The left, middle and right images are drawn from the first, second and third terciles of the distances to the cluster centroids.}
    \label{fig:images_cluster}
\end{figure}

Figure \ref{fig:images_cluster} shows three images for each of the clusters containing at least 20 protest images in the test set. The left images are drawn from the first tercile, the middle images from the second tercile and the right images from the third tercile of each cluster.

\clearpage

\section{Results using Secondary Dataset}
\label{app:results-secondary}

The paper's primary dataset uses high and low certainty protest images as protest images and high and low certainty non-protest images as non-protest images. This coarsening may introduce noise, so we repeat the analysis with a secondary dataset using only the high confidence protest and non-protest images.

\begin{figure}[h!]
    \centering
    \includegraphics[width=0.96\textwidth]{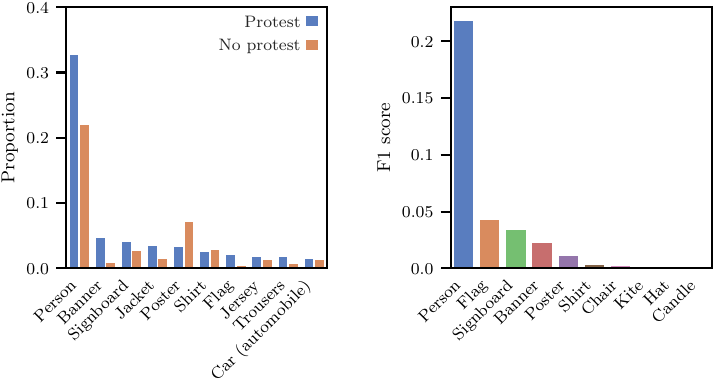}
    \caption{Proportion of segments on high confidence protest and non-protest images (left) and importance of area-sum aggregated segments (right) on images that have been annotated with high confidence.} 
    \label{fig:features_similarities_high}
\end{figure}

\begin{figure}[h!]
    \centering
    \includegraphics[width=0.96\textwidth]{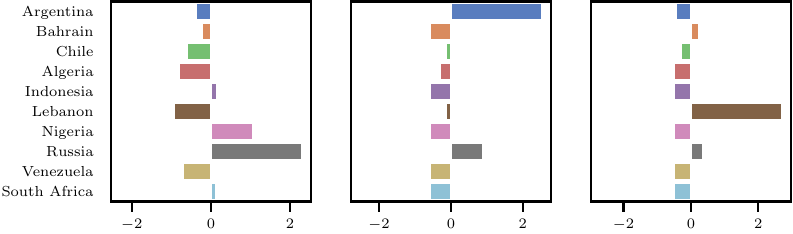}
    \caption{Differences in importance in different countries of posters (left), cars (middle), and candles (right) on images that have been annotated with high confidence.} 
    \label{fig:features_differences_high}
\end{figure}

Table \ref{tab:models_high} shows the fit statistics for the resulting models. All model fits improve and the rank ordering does not change. Figure \ref{fig:features_similarities_high} shows the object categories occupying the largest areas of protest images and the important objects of protest images. Importance results are largely the same, though chairs take kites' place and cars drop out for hats. Figure \ref{fig:features_differences_high} shows the variation of object importance by country. The results for posters, cars and candles are the same.

\begin{table}[h!]
\centering
\small
\begin{tabular}{lrrrrrr}
\toprule
 & \multicolumn{3}{c}{Training} & \multicolumn{3}{c}{Testing} \\
 & Precision & Recall & F1 & Precision & Recall & F1 \\
\midrule
Segments (COCO, bin, logistic) & 0.6622 & 0.3143 & 0.4263 & 0.6309 & 0.2986 & 0.4054 \\
Segments (COCO, count, logistic) & 0.6555 & 0.4712 & 0.5483 & 0.6607 & 0.4768 & 0.5539 \\
Segments (COCO, area max, logistic) & 0.1241 & 0.0018 & 0.0036 & 0.0690 & 0.0009 & 0.0017 \\
Segments (COCO, area sum, logistic) & 0.1629 & 0.0039 & 0.0076 & 0.1698 & 0.0039 & 0.0076 \\
\midrule
Segments (COCO, bin, tree) & 0.7797 & 0.3930 & 0.5226 & 0.5602 & 0.2883 & 0.3807 \\
Segments (COCO, count, tree) & 0.5233 & 0.7011 & 0.5992 & 0.5296 & 0.7009 & 0.6033 \\
Segments (COCO, area max, tree) & 0.7881 & 0.5979 & 0.6799 & 0.5129 & 0.3941 & 0.4457 \\
Segments (COCO, area sum, tree) & 0.7707 & 0.6184 & 0.6862 & 0.5235 & 0.4225 & 0.4676 \\
\midrule
Segments (COCO, bin, forest) & 0.6382 & 0.2766 & 0.3859 & 0.5707 & 0.2517 & 0.3494 \\
Segments (COCO, count, forest) & 0.8051 & 0.5486 & 0.6525 & 0.7105 & 0.4901 & 0.5801 \\
Segments (COCO, area max, forest) & 0.8525 & 0.3496 & 0.4959 & 0.6754 & 0.2435 & 0.3580 \\
Segments (COCO, area sum, forest) & 0.8685 & 0.3759 & 0.5247 & 0.6752 & 0.2612 & 0.3767 \\
\midrule
Segments (COCO, bin, boosted tree) & 0.7558 & 0.4305 & 0.5486 & 0.6507 & 0.3559 & 0.4601 \\
Segments (COCO, count, boosted tree) & 0.7439 & 0.6084 & 0.6694 & 0.6999 & 0.5800 & 0.6344 \\
Segments (COCO, area max, boosted tree) & 0.9373 & 0.7413 & 0.8279 & 0.6504 & 0.4290 & 0.5170 \\
Segments (COCO, area sum, boosted tree) & 0.7827 & 0.5154 & 0.6216 & 0.6700 & 0.4281 & 0.5224 \\
\midrule
Segments (LVIS, bin, logistic) & 0.7813 & 0.6290 & 0.6969 & 0.7705 & 0.6183 & 0.6861 \\
Segments (LVIS, count, logistic) & 0.7526 & 0.5402 & 0.6290 & 0.7312 & 0.5301 & 0.6146 \\
Segments (LVIS, area max, logistic) & 0.4146 & 0.0423 & 0.0768 & 0.4017 & 0.0396 & 0.0721 \\
Segments (LVIS, area sum, logistic) & 0.5257 & 0.1014 & 0.1700 & 0.5588 & 0.1063 & 0.1786 \\
\midrule
Segments (LVIS, bin, tree) & 0.7628 & 0.5247 & 0.6217 & 0.7119 & 0.4806 & 0.5739 \\
Segments (LVIS, count, tree) & 0.7906 & 0.5621 & 0.6571 & 0.7477 & 0.5151 & 0.6099 \\
Segments (LVIS, area max, tree) & 0.7880 & 0.5809 & 0.6688 & 0.7228 & 0.5262 & 0.6091 \\
Segments (LVIS, area sum, tree) & 0.7865 & 0.5819 & 0.6689 & 0.7208 & 0.5275 & 0.6092 \\
\midrule
Segments (LVIS, bin, forest) & 0.9257 & 0.4746 & 0.6274 & 0.8321 & 0.3881 & 0.5293 \\
Segments (LVIS, count, forest) & 0.9488 & 0.5926 & 0.7295 & 0.8113 & 0.4514 & 0.5800 \\
Segments (LVIS, area max, forest) & 0.9681 & 0.5842 & 0.7287 & 0.8265 & 0.4139 & 0.5516 \\
Segments (LVIS, area sum, forest) & 0.9878 & 0.5815 & 0.7321 & 0.8874 & 0.4002 & 0.5516 \\
\midrule
Segments (LVIS, bin, boosted tree) & 0.9977 & 0.9877 & 0.9927 & 0.7950 & 0.6639 & 0.7236 \\
Segments (LVIS, count, boosted tree) & 0.9906 & 0.9550 & 0.9725 & 0.8124 & 0.6876 & 0.7448 \\
Segments (LVIS, area max, boosted tree) & 1.0000 & 1.0000 & 1.0000 & 0.8325 & 0.6863 & 0.7524 \\
\fontseries{b}\selectfont Segments (LVIS, area sum, boosted tree) & \fontseries{b}\selectfont 1.0000 & \fontseries{b}\selectfont 1.0000 & \fontseries{b}\selectfont 1.0000 & \fontseries{b}\selectfont 0.8301 & \fontseries{b}\selectfont 0.7001 & \fontseries{b}\selectfont 0.7596 \\
\midrule
ResNet50 (Won et al., 2017) & 0.5686 & 0.5305 & 0.5489 & 0.5694 & 0.5245 & 0.5460 \\
ResNet50 (self-trained) & 0.8987 & 0.8428 & 0.8698 & 0.8251 & 0.7612 & 0.7919 \\
\fontseries{b}\selectfont ViT (self-trained) & \fontseries{b}\selectfont 0.9516 & \fontseries{b}\selectfont 0.9271 & \fontseries{b}\selectfont 0.9392 & \fontseries{b}\selectfont 0.8917 & \fontseries{b}\selectfont 0.8649 & \fontseries{b}\selectfont 0.8781 \\
\bottomrule
\end{tabular}
\caption[Evaluation of different methods with high confidence]{Evaluation of different methods on images that have been annotated with high confidence. ``Self-trained'' means trained on the images collected for this project.}
\label{tab:models_high}
\end{table}

\clearpage

\section{Temporal Analysis} 
\label{app:temporal}

We analyze how the prevalence of the segments changes over the course of a protest. Having collected the images in our dataset based on protest periods in countries, we can track their prevalence before, during and after the protests. To do this, we use the LVIS segments that we detected on the images in our dataset. From these segments, we sum up the occurrence of segments for each country, day and segment type.

\begin{figure}[h!]
\centering
\includegraphics[width=0.85\textwidth]{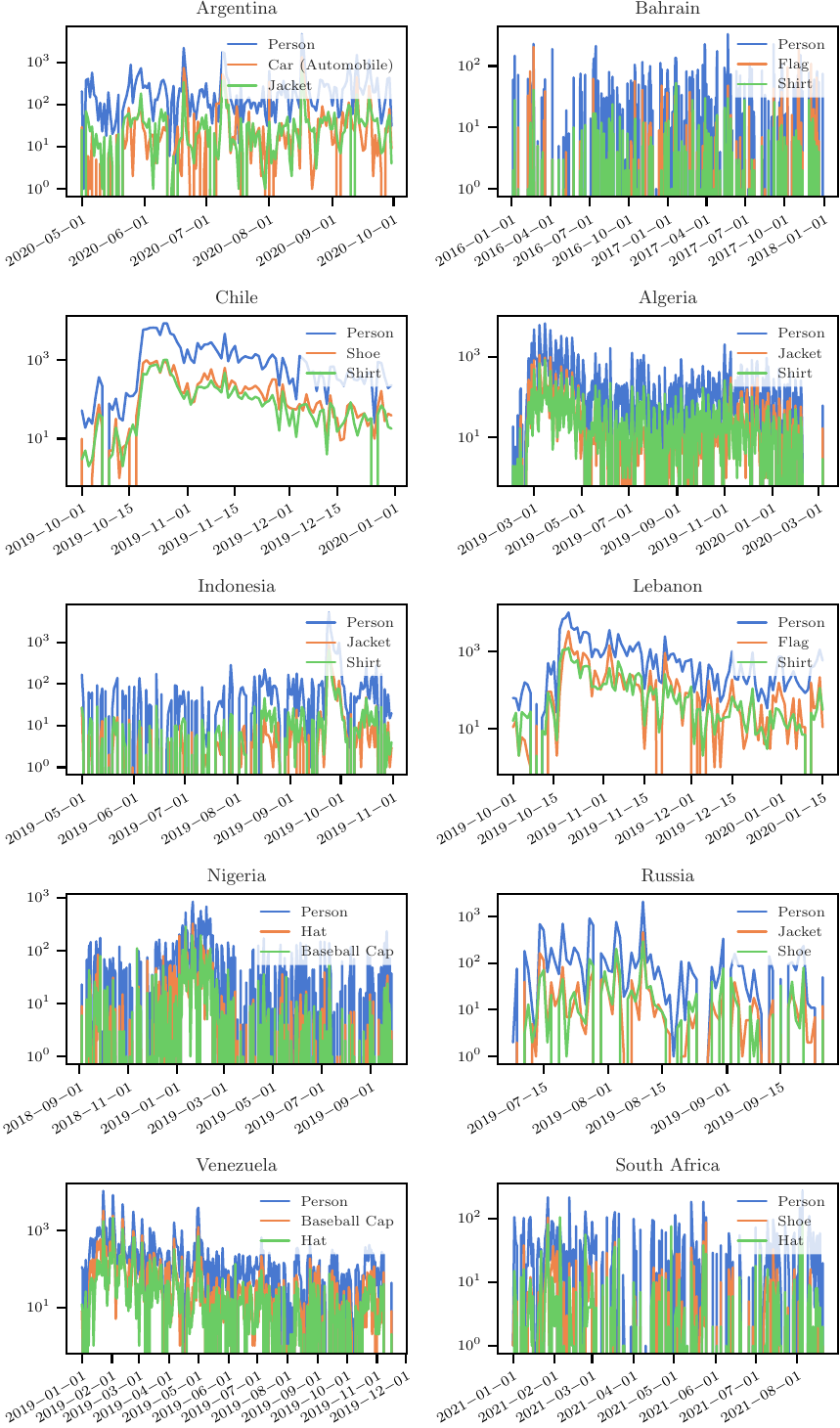}
\caption{The three most common segments per country and their frequency over time.}
\label{fig:dynamics}
\end{figure}

Figure \ref{fig:dynamics} displays the frequency of the segments over time. In each country, it is limited to the three most frequent segments that are detected in that country over the entire period.

\clearpage

\section{Analysis of Segments}
\label{app:segments}

Our main analysis in the paper shows which object categories are identified by a machine to be important for recognizing a protest image. However, are these categories also considered to be important by humans? To find out, we conduct an additional validation exercise at the level of segments (not entire images), asking a coder which segments they deem important for recognizing a protest image. 

For this task, we create a subsample of our protest image dataset consisting of 100 random protest images (high or low confidence) from each of the 10 countries. These images are inspected by one of our coders, who then have to complete two coding steps. In the first step, the coder is asked to look at the protest image and name up to three objects that the coder considers most important to identify it as a protest image. The identification of these objects is done on the raw images. In the second step, after the objects have been freely named, the same protest image is shown with the segments highlighted. The segments shown are those from the LVIS vocabulary \citep{gupta2019lvis} that were recognized by the segmentation model by \citet{zhou2022detecting}, with a confidence score of at least 0.1 (as for the analysis in the paper). Importantly, the segment categories are not shown for these segments, they are simply numbered. The coder is then asked whether the objects identified as important in the first step correspond to one of the segments shown. Because some of the images contain a large number of segments, making it difficult to find the correct identifiers, the coding tool is configured such that the coder could interactively click through the segments to find the right segments and numbers.

The coder identifies 2,210 objects as important on the 1,000 protest images. The ten most frequent object names (freely chosen by the coder) are: people (776), flag (430), poster (216), signboard (195), banner (118), mask (97), police (84), person (69), fire (43) and kid (15). These categories largely overlap with those from our two-stage classification method, which identifies people as the most important objects, followed by flags, signboards, banners and posters. Our method does not identify police officers, fires and children because they are not included in the LVIS vocabulary as separate categories. This shows that custom adaptations of the segmenting method for specific tasks will likely improve results, as we discuss in the paper. A first result is that at a general level, the object categories identified by our two-stage method largely match those that coders consider relevant for the identification of protest images.  

We also test whether \textit{all} segments identified by the coder could be matched to LVIS segments. For the vast majority, this is possible. It is only for 141 objects (6.4\%) that there is no corresponding segment. These include categories that are included in the LVIS vocabulary (for example, 24 posters, 22 persons, 19 flags), but which the segmenter fails to identify on the respective images. For the successfully assigned objects we check whether the objects indeed match the segments. To do this, we compare the object names given by the coder with the ones detected by the segmenter. We set up a small dictionary to ensure that object names that are spelled slightly differently could be recognized as identical. For a strict matching (object names identical), we find that 1,512 out of the 2,210 segments (68\%) are correctly detected by the segmenter. For a lenient matching (the dictionary incorporates subtypes and supertypes as well), the number of correctly detected segments increases to 1,626 (74\%). Objects that are repeatedly incorrectly recognized are objects that are interpreted as weapons by our coder. This analysis shows that human and machine largely rely on the same segments to code protest images.


\clearpage

\section{Collecting Images in the Future}
\label{app:images_future}

The data for this paper were collected in real time using R's \texttt{streamR} package \citep{barbera2018streamr}. One of the authors maintained a continuous connection to Twitter's filtered stream endpoint and requested only tweets with location information. Collecting tweets agnostically in real-time means any event of sufficient magnitude is collected automatically, obviating the need for researchers' to search for events \emph{post hoc}. Searching for events after they occur also risks introducing sample bias, as searches rely on keywords or specifying users and content could be deleted between its posting and the researcher's download.  

The past tense is used in the previous paragraph because Elon Musk's purchase of Twitter has led to severely restricted data access.  The free tier only returns 1,500 tweets per month. The Basic tier for \$100/month provides only 10,000 tweets. The Pro, \$5,000 and 1 million. At the Pro level, one can stream tweets, but they count against the 1 million quota, which would be reached in less than a day without stringent filter rules. Access equivalent to what this paper had requires the Enterprise tier. That pricing is available upon request in contrast to the \$0 price before Musk's neutering. Except for three alternatives or academics with corporation-level resources, Twitter access is over.

The three alternatives are using already downloaded tweets, scraping, and the European Union's Digital Services Act (DSA). If one has previously downloaded tweets, it is possible and easy to download media from those tweets. Each tweet contains a \texttt{image\_url} field, and access to those media are not rate limited. While old tweets may no longer be available \citep{pfeffer2023this}, if they are then their images are.    
As of April 2022, \emph{hiQ v. LinkedIn} and then \emph{Van Buren v. United States}, decided at the United States' Ninth Circuit Court of Appeals and the Surpreme Court, respectively, establish that the Computer Fraud and Abuse Act does not allow companies to prevent scraping of their public data. A researcher can build a scraper themselves or use the Python package \texttt{snscape}. The other option is to apply for research access as permitted pursuant to Article 40 of the DSA. Doing so requires an application. As of this writing, we are not aware of any Twitter research conducted as a result of the DSA.

This paper's method is applicable to any image data, however, not just those from Twitter. Other sources of images include Facebook pages, Instagram, Telegram, and WhatsApp; Pexels, Tumblr, and Unsplash; news archives; or stills of videos from TikTok and YouTube. A golden age of online social media data appears to have ended, but researcher creativity will ensure that research does not.

\clearpage

\printbibliography